\documentclass[11pt]{article}
\usepackage[utf8]{inputenc}
\usepackage{amsmath,amsthm,amssymb,mathtools,array,bbm,float,bm}
\usepackage{subcaption} 
\usepackage{authblk}
\usepackage{multirow}
\usepackage{color}
\usepackage{url}
\usepackage{pdflscape}
\usepackage{enumitem}
\usepackage[margin=1in]{geometry}
\usepackage[utf8]{inputenc}

\allowdisplaybreaks[1]

\usepackage{xr}
\makeatletter
\newcommand*{\addFileDependency}[1]{
  \typeout{(#1)}
  \@addtofilelist{#1}
  \IfFileExists{#1}{}{\typeout{No file #1.}}
}
\makeatother

\title{On Aligning Prediction Models with Clinical Experiential Learning: A Prostate Cancer Case Study}
\author[1]{Jacqueline J. Vallon, MS}
\author[1]{William Overman, MS}
\author[1]{Wanqiao Xu, BS}
\author[2]{Neil Panjwani, MD}
\author[1]{Xi Ling, BS}
\author[1]{Sushmita Vij, PhD}
\author[1]{Hilary P. Bagshaw, MD}
\author[1, 3]{John T. Leppert, MD, MS}
\author[1]{Sumit Shah, MD, MPH}
\author[1]{Geoffrey Sonn, MD}
\author[1]{Sandy Srinivas, MD}
\author[1, 3]{Erqi Pollom, MD, MS}
\author[1*]{Mark K. Buyyounouski, MD, MS}
\author[1*]{Mohsen Bayati, PhD}
\affil[1]{\normalsize{Stanford University, Stanford, CA}}
\affil[2]{\normalsize{University of Washington, Seattle, WA}}
\affil[3]{\normalsize{Palo Alto Veterans Affairs Hospital, Palo Alto, CA}}
\affil[*]{\normalsize{Served as equally contributing co-senior authors}}

\date{}

\begin{document}

\maketitle


\begin{abstract}
    Over the past decade, the use of machine learning (ML) models in healthcare applications has rapidly increased. Despite high performance, modern ML models do not always capture patterns the end user requires. For example, a model may predict a non-monotonically decreasing relationship between cancer stage and survival, keeping all other features fixed. In this paper, we present a reproducible framework for investigating this misalignment between model behavior and clinical experiential learning, focusing on the effects of underspecification of modern ML pipelines. In a prostate cancer outcome prediction case study, we first identify and address these inconsistencies by incorporating clinical knowledge, collected by a survey, via constraints into the ML model, and subsequently analyze the impact on model performance and behavior across degrees of underspecification. The approach shows that aligning the ML model with clinical experiential learning is possible without compromising performance. Motivated by recent literature in generative AI, we further examine the feasibility of a feedback-driven alignment approach in non-generative AI clinical risk prediction models through a randomized experiment with clinicians. Our findings illustrate that, by eliciting clinicians' model preferences using our proposed methodology, the larger the difference in how the constrained and unconstrained models make predictions for a patient, the more apparent the difference is in clinical interpretation. 
\end{abstract}


\section{Introduction}
Over the past decade, given the advancement in artificial intelligence (AI), there has been an increasing shift toward utilizing data-driven approaches, such as machine learning (ML) models, to support decision making in healthcare applications. For example, ML techniques are used for clinical risk prediction modeling across healthcare domains, including for cancer prognosis \cite{kourou_machine_2015}, septic shock prediction \cite{henry_targeted_2015}, and detection of inpatient deterioration \cite{escobar_piloting_2016}, to name a few. However, healthcare is a field that is deeply rooted in clinical experiential learning, with experts gaining valuable in-depth knowledge as they progress throughout their career that cannot always be explained by data. The recent reliance on data-driven approaches has therefore introduced a new challenge: a fully data-driven approach is efficient and can detect new knowledge found in the data, but has weaknesses, such as being prone to (incorrectly) learning anomalies in the data and likely being underspecified; in contrast, a clinician-driven approach is trusted in practice and captures necessary, known clinical observations and behaviors to drive decisions, but may miss knowledge that requires processing large volumes of data. In this paper, despite the potential benefits of processing large volumes of data, we posit that model alignment with foundational clinical experiential learning is crucial to building trustworthy models, and with this perspective, showcase potential pitfalls of a fully data-driven approach and illustrate methods to incorporate clinical knowledge to mitigate these.

\begin{figure}[H]
 \centering
 \begin{subfigure}[b]{0.49\textwidth} 
    \centering
   \includegraphics[width=\linewidth]{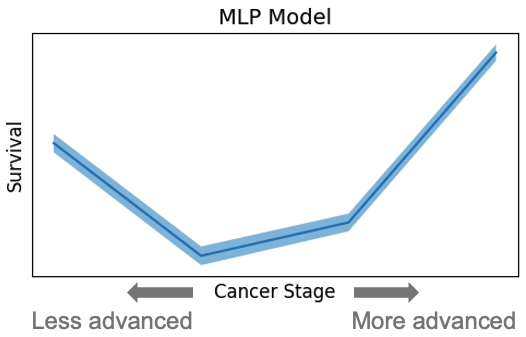}
   \caption{MLP Model}
   \label{subfig-1:mlp}
 \end{subfigure}%
 \hfill
 \begin{subfigure}[b]{0.49\textwidth}
    \centering
   \includegraphics[width=\linewidth]{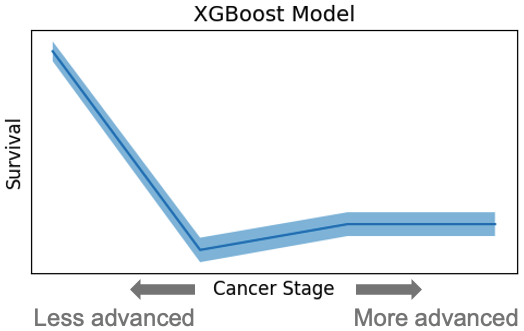}
   \caption{XGBoost Model}
   \label{subfig-2:xgboost}
 \end{subfigure}
 \caption{Examples of behavior inconsistent with clinical experiential learning in an MLP and XGBoost model.}
 \label{fig:motivating_examples}
\end{figure}

Regardless of the high performance of ML models and even when trained with sufficient amounts of data, there is growing evidence in the literature that complex, modern ML models do not always effectively capture patterns that are required by the end user when deployed in practice \cite{wang_deontological_2020}. To illustrate an example of this inconsistency, consider the task of predicting survival post-treatment for patients diagnosed with prostate cancer. Figure \ref{fig:motivating_examples} shows the predicted relationship between cancer stage and survival for two ML models: a multilayer perceptron (MLP) model (left) and an Extreme Gradient Boosting (XGBoost) model \cite{chen_xgboost_2016} (right). Clinical experiential learning, which captures the fundamental collective knowledge of clinicians gained through extensive clinical observations and experiences, anticipates a non-increasing relationship between cancer stage and survival, setting all other features fixed. However, both models display a behavior that is inconsistent with this prerequisite, which can curtail the model's adoption in clinical practice. Moreover, these models describe very different behaviors despite being trained on the same data. 

Identifying the exact causes behind these inconsistencies (i.e., model behavior inconsistent with requirements by the end user) can be challenging. Potential factors include data selection bias or noise bias. Another contributor that can enable these inconsistencies in modern ML pipelines is the \textit{underspecification} of these pipelines, meaning many ML models with equivalently strong held-out performance can be produced by the same pipeline \cite{d'amour}. While recent literature in the theory of deep learning points to the importance of overparameterization and its implicit regularization to help improve the test error of models (i.e., a form of benign underspecification) \cite{belkin_reconciling_2019, mei_generalization_2022, nakkiran_deep_2021}, a practical implication of underspecification is that models can have strong performance, but be `incorrect' for practical reasons.

This raises the following questions for supervised clinical risk prediction ML models:

\begin{enumerate}
    \item Why do these inconsistencies occur in modern ML pipelines? What role does underspecification play?
    \item How can one identify and address these inconsistencies to better align an ML model with clinical experiential learning? What is the impact of such alignment on model performance and behavior across degrees of underspecification?
    \item Can clinicians detect differences between aligned and misaligned models?
\end{enumerate}

In this work, we study these research questions in supervised risk prediction ML models applied to a clinical case study: prostate cancer outcome prediction. We begin by focusing on how one can identify these inconsistencies and subsequently align the models with clinical experiential learning. We then analyze the consequences of the alignment on model performance and behavior, and on clinical interpretation. 

In the case study, we focus on clinically expected monotonicity, meaning as the feature value increases, the output of the function should be always non-decreasing or always non-increasing, \textit{keeping all other features fixed} \cite{groeneboom_nonparametric_2014}. With this perspective, we define a model behavior to be \textit{inconsistent} with clinical experiential learning if the model does not uphold the expected monotonicity in select features (i.e., the model violates monotonicity in select univariate partial dependence plots). To first identify and then resolve inconsistencies, we take a \textit{constraint-guided alignment approach via knowledge elicitation}. To identify the inconsistencies, we need clinician input and, therefore, directly elicit from clinicians which features have an expected monotonicity. To resolve the inconsistencies, we then encode this knowledge as constraints into the training of the ML model using existing shape-constrained methods. We subsequently compare the performance and behavior of a constrained and unconstrained model across degrees of underspecification by varying the train set size. In the case study, we use XGBoost models given their high performance in applications, they are non-parametric, and they allow for integrating constraints into the model.\footnote{A simpler, preliminary analysis that uses a gradient boosted machine model to show an example of inconsistencies in the clinical setting of prostate cancer outcome prediction (similar to Figure \ref{fig:motivating_examples} in this paper) and collects and incorporates survey results to produce a constrained model applied to a LASSO model (similar to Sections \ref{sec:methods-solution-approach} and \ref{sec:results-univariate}) is described in an extended abstract published by a subset of the authors of this paper in the \textit{International Journal of Radiation Oncology, Biology, Physics} \cite{astro_abstract}.}

Inspired by the results of the constraint-guided alignment approach and recent advances in generative AI models \cite{ouyang_training_2022}, we further leverage the models built in the constraint-guided alignment approach to examine whether a key requirement for a potential \textit{feedback-driven alignment approach} can be met in non-generative AI models. At a high level, a feedback-driven alignment approach would work as follows: first, clinicians are shown two model predictions for a single patient and are asked to provide their preference; second, this step is repeated for many patients; and finally, this clinical feedback is incorporated as a reward signal to improve model behavior. A key requirement and practical challenge of this approach is that clinicians need to be able to detect differences between two models when assessing patient-level predictions and determine which is better aligned with clinical experiential learning. This step is usually relatively trivial in generative AI models but can be more complex in non-generative AI models (e.g., classical risk prediction ML models), where the models normally just make a single numerical prediction, which can be hard to evaluate and compare across different models. Therefore, this approach requires an alternative and clinician-interpretable kind of output that captures how models make predictions. To understand whether this key requirement for a feedback-driven alignment approach can be fulfilled in non-generative AI models, we build on our work from the constraint-guided alignment approach and perform an exploratory experiment. In this experiment, we show clinicians select information on how constrained and unconstrained models make predictions at a patient-level, and analyze whether and when clinicians can detect differences between these models. While models used in a feedback-driven alignment approach are not limited to constrained and unconstrained models, we intentionally use these models in this chapter, because we know that the constrained model is better aligned with clinical experiential learning.


\section{Related Work}
Our work is related to four areas of literature (non-mutually exclusive): the underspecification of modern ML pipelines; shape constraints and isotonic regression; a feedback-driven alignment approach in generative AI models; and domain knowledge informed ML models. While it is most closely related to the literature on underspecification, it combines methods of the others as alignment techniques, to study the relationship between underspecification and the impact of violations of clinically expected requirements in a clinical risk prediction model.

%

\subsection{Underspecification of ML Pipelines}
As ML models become more complex, they have greater freedom in the behaviors they can learn. Recent literature draws attention to the consequences of this, particularly concerned with the disparity between a model's test performance generalizability and its practical implications \cite{d'amour}. Consequently, there is a growing need to broaden our evaluation criteria for modern ML models through `stress tests' that are designed to study underspecification, and to inspect models beyond just their performance. Common stress tests, outlined by D'Amour et al. \cite{d'amour}, include: subgroup analyses, which examine the model's behavior across subsets of the data; distribution shift evaluations, which assess the model's performance across test distributions; and contrastive evaluations, which analyze whether the model's output exhibits inconsistent behavior with specific input modifications. Such stress tests have been applied in a variety of applications to illustrate the consequences of underspecification, including medical imaging \cite{eche_toward_2021, d'amour}, natural language inference \cite{mccoy_right_2019, d'amour}, clinical predictions \cite{tang_identifying_2022, d'amour}, and many more. In their work, Hinns et al. \cite{hinns_initial_2021} propose another method, using feature attribution techniques, to detect underspecification, by examining the different explanations generated by ML models produced by the same pipeline. Results of these stress tests have given rise to a need to develop methods that reduce the impact of underspecification, especially when the predicted model does not adhere to behaviors essential for a given application. Some proposed strategies include: imposing constraints on the type of model that can be learned \cite{teney_predicting_2022, ross_right_2017} or augmenting the train set \cite{mccoy_right_2019}. 

In this work, we study a different type of model behavior, primarily the violation of monotonicity in univariate partial dependence plots, in a clinical risk prediction modeling case, to understand the potential implications of underspecification. To do so, we take a unique approach by analyzing the impact of incorporating clinical experiential learning to improve model alignment with the end user across degrees of underspecification.

\subsection{Shape Constraints and Isotonic Regression}
Monotonic behavior is relevant to many applications. In past literature, monotonic constraints are integrated into ML models for a variety of reasons, including: 1) data-related purposes \cite{altendorf_learning_2005, velikova_monotone_2006}, for example, to reduce the hypothesis space in the case of sparse or noisy training data; 2) model-related considerations \cite{dugas_incorporating_2009, potharst_classification_2002, velikova_monotone_2006, wang_deontological_2020}, for example, to prevent overfitting in overparameterized nonlinear models, to increase model stability, or to modify models that do not typically exhibit monotonic behavior despite being trained on monotonic data; and 3) user-related motivations \cite{velikova_monotone_2006, pazzani_acceptance_2001}, for example, to enhance the decision-making process and establish user trust by aligning the model's behavior with the decision makers' expertise. Literature on shape constraints focuses on two main aspects: first, the integration of shape constraints into a range of ML models \cite{cano_monotonic_2019}, including decision trees, ensembles, and deep learning models like neural networks \cite{runje2023constrained}; and second, the application of such techniques in domains such as credit rating \cite{chen_credit_2014}, house pricing prediction \cite{potharst_classification_2002}, and even to promote fairness by correcting violations against social norms \cite{wang_deontological_2020}. 

In our work, while we impose existing shape-constrained methods to correct inconsistencies with clinical experiential learning in modern ML models, our primary focus is to analyze the occurrence of the inconsistencies and the impact of their correction through shape constraints in domains of varying degrees of underspecification, which, to the best of our knowledge, has not been studied before.

\subsection{Feedback-Driven Alignment Approach in Generative AI Models}
Model alignment through human feedback has accelerated the adoption of large language models, like ChatGPT, in the past couple of years and is a very active area of research for generative AI models. The paper by Ouyang et al. \cite{ouyang_training_2022} outlines an approach to train language models with human feedback and illustrates its value. A key requirement to this approach is that an end user can rank outputs of a model from best to worst, an often trivial step in generative AI models, where the differentiation is between two sentences or images. 

In this work, we ask ourselves whether there is a future opportunity to implement such a feedback-driven alignment approach in \textit{non-generative AI models}. Adapted from the approach by Ouyang et al. \cite{ouyang_training_2022}, one could imagine creating a similar alignment method consisting of four main steps: 1) for a patient, several outputs of an ML pipeline are generated; 2) a clinician then ranks these outputs from best to worst; 3) a reward model is trained on this data; and 4) the pipeline is optimized using the reward model to create a model that is more aligned with clinical experiential learning. In contrast to generative AI models, step 2 can be more complex in classical risk prediction models, where the underlying behavior of the model must be assessed rather than just the single numerical outcome. The primary goal of our analysis is to leverage the models from the constraint-guided alignment approach to analyze whether step 2 is feasible through our exploratory experiment.

\subsection{Domain Knowledge Informed ML Models}
Overall, our work fits into the broader literature on incorporating domain knowledge into ML models to increase their robustness and adoption in practice. This literature dates back several decades in the medical domain: for example, Pazzani et al. \cite{pazzani_acceptance_2001} illustrate an example in the early 2000s in which experts prefer rule learning systems consistent with medical knowledge by adding monotonic constraints, with no adverse impact on the performance of the rules. Our work extends these ideas to modern ML pipelines. More recently, Borghesi et al. \cite{borghesi_improving_2020} provide an overview of literature that explores the consequences of injecting domain knowledge into different parts of the ML pipeline, particularly focused on deep learning models, with broader generalizability. They mention the use of domain knowledge to modify the hypothesis space of the model to restrict the types of structures the model can learn. A recent illustration of this in the clinical application of prostate cancer is provided by Elmarakeby et al. \cite{elmarakeby_biologically_2021}. In their work, they develop a biologically informed deep learning model that classifies prostate cancer patients by their treatment-resistance states, resulting in greater model interpretability by reducing arbitrary overparameterization. 

This article leverages domain knowledge to explore why inconsistencies with clinical experiential learning occur in modern, underspecified ML pipelines and how to mitigate them. 

\section{Case Study: Methods}

\subsection{Overview of Clinical Case Study}
For the case study, we use data from a de-identified National Cancer Database (NCDB) file \cite{ncdb}. The NCDB, founded in 1989, is a nationwide, facility-based, comprehensive clinical surveillance resource oncology database created by the American Cancer Society and the Commission on Cancer of the American College of Surgeons. While we restrict our analyses to a subset of the years in the data where the coverage may have been different, the NCDB currently includes 72\% \cite{mallin_incident_2019} of all newly diagnosed cancer cases in the US each year. We specifically focus on a cohort of patients who were diagnosed with and treated for prostate cancer.\footnote{Both the American College of Surgeons and the Commission on Cancer have not authenticated nor taken responsibility for the methodologies implemented in this study or the conclusions developed. The American College of Surgeons has an established Business Associate Agreement with each of its Commission on Cancer accredited hospitals, including a data use agreement.} This study is IRB exempt.

\subsubsection{Patient Cohort and Outcome Definition}\label{sec:clinical-consistency-patient-cohort}
For our patient cohort ($n=85,432$), we are interested in patients treated for prostate cancer with a prostatectomy or radiation therapy (RT) for their first course of treatment (note: all treatment information available in the data relates to a patient's first course of treatment). We consider a patient as treated with a prostatectomy if it is indicated that the surgical procedure performed to the primary site at any facility is classified as a prostatectomy. We consider a patient as treated with RT if they received internal radiation (i.e., radioactive implants), external radiation (i.e., beam radiation), or a combination radiation (i.e., beam radiation with radioactive implants or radioisotopes) to the prostate and/or pelvis. 

To identify the patient cohort, we first apply a set of filters, based on recommendations provided in the NCDB user files and our team's discussions of the patient cohort of interest. Appendix \ref{sec:appendix-patient-cohort} details this first set of filters that we apply. Broadly, these filters ensure that we can identify patients with the above-outlined treatments, calculate our outcome of interest for each patient, and remove cases that are likely to have data inputs that are incomplete or not standardized across patients. 

Once the data is narrowed down using these filters, we compute the outcome of interest for each patient, applying additional filters required along the way to accurately classify patients given the data available. For the outcome, we consider the binary outcome of 10-year survival post-first course of treatment (prostatectomy or type of radiation therapy). To compute this, for each patient, we first determine the time between the end of their treatment and their date of last contact (at which point they are known to either be alive or have died). We then remove all patients who are last contacted less than 10 years post-treatment and were noted to be alive at that time in the dataset (implying that we do not have access to enough follow-up information about them to determine their 10-year survival). We then classify the remaining patients as `alive' at 10 years post-treatment if \textit{either:} 1) their last contact was at least 10 years post-treatment during which they were noted to be alive in the dataset, \textit{or} 2) their last contact was more than 10years post-treatment during which they were noted to have died in the dataset; and we classify a patient as `dead' 10 years post-treatment if their last contact was at most 10 years post-treatment and they were noted to have died in the dataset. Given the 10-year follow-up requirement to calculate this outcome, we limit our analyses to patients diagnosed in the years 2004-2005. In our final patient cohort, 74.9\% of patients are alive 10 years post-treatment.



Table \ref{tab:patient_cohort} lists summary statistics for our final patient cohort. We recognize that the above filters may create some selection bias, but we do not expect such biases to significantly impact the results of our main objective, which is to study the misalignment of ML model behavior with clinical experiential learning and the impact of underspecification, rather than to build a model to deploy in practice. We did complete some distributional analyses for key features, which compare distributions of features in our final dataset to the distributions in the original dataset (i.e., before any filters are imposed) (see Appendix \ref{sec:appendix-patient-cohort}). While we do not seem to observe too large of changes for several of the summary statistics, the selected cohort generally seems to consist of a healthier population.

\begin{table}[h]
\caption{Summary statistics of patient cohort.}
\centerline{\begin{tabular}{  l l l } 
    \cline{2-3} & \\[-1.5ex]
     & \multicolumn{2}{l}{\textbf{Metric for Numerical Features}} \vspace{3pt} \\
    \cline{2-3} & \\[-1.5ex]
    \textbf{n=85,432} & \textbf{Mean (SD)} & \textbf{Median} \vspace{3pt}\\ 
    \hline\hline & \\[-1.5ex]
    Age at Diagnosis & 64.7 (8.5) & 65.0 \vspace{3pt}\\ 
    Clinical Stage Group & 2.0 (0.3) & 2.0 \vspace{3pt}\\
    Comorbidity Score & 0.15 (0.4) & 0.0 \vspace{3pt}\\
    Highest Pre-Treatment PSA & 9.9 (13.3) & 6.0 \vspace{3pt}\\
    Primary Gleason Pattern & 3.2 (0.5) & 3.0 \vspace{3pt}\\
    Secondary Gleason Pattern & 3.4 (0.6) & 3.0 \vspace{3pt}\\
    \hline & \\[-1.5ex]
    Treatment Type &  & \vspace{3pt}\\
    \hspace{0.5cm} Prostatectomy & 48.6\% & \\
    \hspace{0.5cm} Internal RT & 17.0\% & \\
    \hspace{0.5cm} External RT & 27.7\% & \\
    \hspace{0.5cm} Combination RT & 6.6\% & \vspace{3pt}\\
    Race = White & 83.5\% & \vspace{3pt}\\
    Insurance Status & & \vspace{3pt}\\
    \hspace{0.5cm} Medicaid & 1.3\% & \\
    \hspace{0.5cm} Medicare & 45.2\% & \\
    \hspace{0.5cm} Other Gov & 1.3\% & \\
    \hspace{0.5cm} Private & 48.3\% & \\
    \hspace{0.5cm} Uninsured & 1.1\% & \vspace{3pt}\\
    \hline\hline
\end{tabular}}
\label{tab:patient_cohort}
\end{table}

\subsubsection{Features}
To predict the outcome, we control for 23 features (listed in Table \ref{tab:features}), including demographic and socioeconomic features, pre-treatment clinical and cancer-specific features, and treatment-specific features.

For data pre-processing, we one-hot encode categorical features (treatment type, facility type, facility location, race, Spanish or Hispanic origin, and insurance status). We consider all other features to be numerical features, implying an inherent order to their values, which requires changing some feature values to numerical values. For example, clinical T stage, which represents the cancer stage of the original tumor, is clinically noted by one of the following: {\tt cT0, cT1a, cT1b, cT1c, cT2a, cT2b, cT2c, cT3a, cT3b, cT4}. To encode the fact that there is an order to the clinical T stages, we change these feature values to numerical values of 0-9, corresponding to the 10 ordered stages, respectively. 

In the case study, we use XGBoost models, which can handle missing values. Therefore, in our main results, for numerical features, we keep the missing values, and for categorical features, we create a separate category labeled `nan'. Appendix \ref{sec:appendix-features} outlines the features with missing values and the fraction of time they are missing for each.

\begin{table}[H]
\centering
\caption{List of features included in analyses.}
\label{tab:features}
\begin{tabular}{p{0.33\textwidth}p{0.35\textwidth}p{0.33\textwidth}}
\centering \textbf{\small \underline{Demographic}} 
\small
\begin{itemize}
\setlength\itemsep{-3pt}
\setlength\parsep{-4pt}
\item[$\circ$] Age at Diagnosis
\item[$\circ$] Race
\item[$\circ$] Spanish or Hispanic Origin
\item[$\circ$] Insurance Status
\item[$\circ$] Educational Attainment for Area of Residence
\item[$\circ$] Median Household Income for Area of Residence
\item[$\circ$] Rurality and Urban Influence for Area of Residence
\end{itemize}
& \centering \textbf{\small \underline{Clinical}} 
\small
\begin{itemize}
\setlength\itemsep{-3pt}
\setlength\parsep{-4pt}
\item[$\circ$] Comorbidity Score
\item[$\circ$] Clinical T Stage
\item[$\circ$] Clinical N Stage
\item[$\circ$] Clinical M Stage
\item[$\circ$] Clinical Stage Group
\item[$\circ$] Clinical Prostate Apex \\Involvement
\item[$\circ$] Primary Gleason Pattern
\item[$\circ$] Secondary Gleason Pattern
\item[$\circ$] Highest Pre-Treatment PSA
\end{itemize}
& \centering \textbf{\small \underline{Treatment}}
\small
\begin{itemize}
\setlength\itemsep{-3pt}
\setlength\parsep{-4pt}
\item[$\circ$] Facility Type
\item[$\circ$] Facility Location
\item[$\circ$] Chemotherapy
\item[$\circ$] Hormone Therapy
\item[$\circ$] Immunotherapy
\item[$\circ$] Endocrine Procedures
\item[$\circ$] Treatment Type \\(prostatectomy or RT)
\end{itemize}
\end{tabular}
\end{table}

\subsection{ML Pipeline and Application to Clinical Dataset}\label{sec:methods-ml-pipeline}
For the main outcome prediction ML model, we train gradient boosted decision trees, utilizing the XGBoost algorithm. A XGBoost model trains an ensemble of decision trees and effectively combines these to build a more accurate model. We choose a XGBoost model as the main outcome prediction ML model, because of its high performance across applications, it is non-parametric, and one can seamlessly integrate monotonic constraints into it. While our primary analyses use XGBoost models, it is important to note that the inconsistencies with clinical experiential learning that we analyze in this case study are not limited to XGBoost models and are also observed in other ML models, such as MLP models as seen in the introductory Figure \ref{fig:motivating_examples}.

To create the train and test data, we first split the data into two sets using an 80/20 split, stratifying by the outcome. In our analyses, we vary the size of the train set used to vary the level of underspecification of models. All train sets are drawn randomly from the 80\% split ($n=68,345$) made on the data. For each model, we specify the size of the data used to train the model; for simplicity, we refer to this data as the \textit{train set}, although it varies throughout analyses. In contrast, the 20\% split ($n=17,087$) is used as the \textit{test set}. The test set is purely used to evaluate the models and is not involved in any training of models. 

To train the models, we use a grid search with 5-fold stratified cross-validation (CV) minimizing the area under the receiver operating characteristic curve (AUC-ROC), to tune the learning rate $\{0.01, 0.1, 0.3, 0.5\}$, the number of boosting rounds $\{100, 300, 500\}$, and maximum depth of trees $\{2, 3, 5, 10\}$. After tuning, the model is then retrained on the entire train set with the best parameters found via CV.

To analyze the predicted behavior of models produced by this ML pipeline, we apply the pipeline to the clinical data to predict overall survival 10 years post-treatment and analyze their univariate partial dependence plots. At a high level, these plots show the relationship predicted by the model between a specific feature and the outcome of survival, keeping all other features fixed, thereby providing immediate insight into the type of behavior the model has learned for each feature. Figure \ref{fig:motivating_examples} shows examples of such a plot.

In particular, to create a univariate partial dependence plot for a specific feature, we complete the following steps:
\begin{enumerate}
    \item Compute all non-missing unique values of the feature of interest in the entire dataset, sorted in ascending order. For example, for the primary Gleason pattern, these values would be $[1, 2, 3, 4, 5]$.
    \item For each unique value found in Step 1:
        \begin{enumerate}
        \item Set the feature of interest to the unique value for all patients in the test set. 
        \item Predict the probability of overall survival for all patients in the modified test set using the trained ML model.
        \end{enumerate}
    \item For each unique value, compute the mean predicted survival and standard error of the mean (SE). Then, plot the unique values against the mean predicted survival with a two SE confidence interval (CI) for each value.
\end{enumerate}

\subsection{Implementation of a Constraint-Guided Alignment Approach}\label{sec:methods-solution-approach}
To identify and correct the inconsistencies in an ML model through a \textit{constraint-guided alignment approach}, we elicit required model behaviors directly from clinicians, and thereafter modify the model behavior by imposing monotonic constraints rooted in this clinical experiential learning into the ML pipeline.

\begin{figure}[h]
\centering
\includegraphics[width=0.7\linewidth]{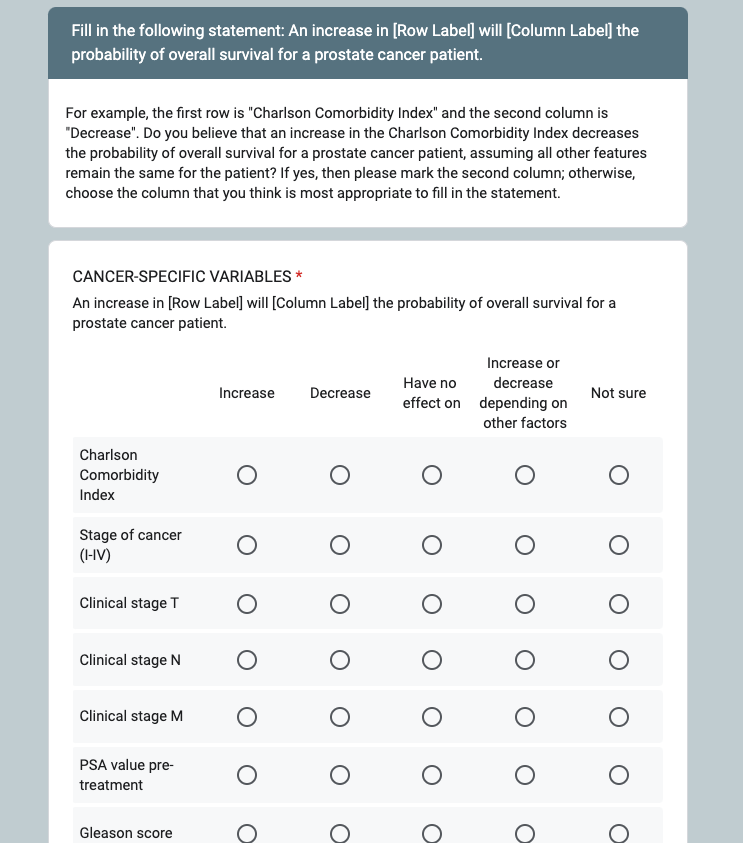}
\caption{Snapshot of the survey sent out to clinicians to collect their clinical intuition.}
\label{fig:query}
\end{figure}

Given the complexity of a clinical setting, the primary effort of this approach lies in the identification of the clinically expected monotonicity in a systematic way. To do this, our team of engineers collaborated closely with a team of clinical experts at the Stanford Cancer Center and the Palo Alto Veterans Affairs Hospital. Together, we first created a library of clinical intuition about the outcome of overall survival post-treatment as related to prostate cancer through a survey. We sent the survey to and received responses from five clinicians. The survey elicits their expertise on what effects an increase in a feature would have on the probability of overall survival for a prostate cancer patient, assuming other features remain fixed. Figure \ref{fig:query} shows an excerpt of the survey.

We are particularly interested in features for which the majority of the clinicians agree that an increase in a feature will either always increase or always decrease the probability of overall survival for a prostate cancer patient. This would signal a required monotonic behavior of that feature in the model, and any behavior that deviates from this is considered inconsistent with clinical experiential learning. Using this logic, we identify the clinical features that should exhibit a monotonic behavior in relation to the outcome of survival, keeping all other features fixed, by narrowing down to features that had such a majority and discussing these further with our main clinical collaborator (M.K.B.). The following features are determined to have a monotonically decreasing relationship with the outcome of survival (total of 9): age at diagnosis, comorbidity score, clinical T stage, clinical N stage, clinical M stage, clinical stage group, highest pre-treatment prostate-specific antigen (PSA) level, primary Gleason pattern, and secondary Gleason pattern. An alternative approach to identifying the clinically expected monotonicity could have been to extensively research published literature to collect information on the relationship between features and the outcome of survival. However, this approach could be very time-consuming and would require that such information, which often stems from clinical observations and experiences, is well-documented for many features. Therefore, our research team decided that the survey was a more efficient and effective approach.

Once these required behaviors are identified, we then align a model with clinical experiential learning by encoding this information as clinically informed monotonic constraints for the above-listed features into the ML pipeline during the training process. To do this, we utilize existing shape-constrained methods that can incorporate monotonic constraints for select features into XGBoost models. 

In the case study, we refer to a trained XGBoost model \textit{with} constraints imposed on it as the `constrained model'. Conversely, we refer to a trained XGBoost model \textit{without} constraints imposed on it as the `unconstrained model'. That is, both models follow the same ML pipeline as described in Section \ref{sec:methods-ml-pipeline}, with the only discrepancy being whether the monotonicity shape constraints are imposed on the model during the training process.

\subsection{Impact of Constraint-Guided Approach Across Degrees of Underspecification}\label{sec:methods-impact}
To study the impact of the constraint-guided approach, we analyze the difference in model performance and behavior of the constrained and unconstrained models across degrees of underspecification, varied by the train set size, with smaller train set sizes being a proxy for greater degrees of underspecification. Specifically, we consider train set sizes in the set \{100, 200, 300, 400, 500, 600, 700, 800, 900, 1000, 1250, 1500, 1750, 2000, 3000, 4000, 6000, 8000\}. For each train set size, we sample 300 different train sets from the 80\% data split  ($n=68,345$) set aside for training purposes, as explained in Section \ref{sec:methods-ml-pipeline}, by varying the random seed used. We then train the constrained and unconstrained models on each train set and compare the below-described performance and distance metrics computed on the held-out test set across train set sizes. We cap the size of the train set considered to 8000 (roughly 10\% of the data), because we generate confidence intervals for these analyses by repeatedly taking random samples from the 80\% data split and want to reduce the correlation across these random sub-samples.

\subsubsection{Impact on Model Performance}\label{sec:methods-impact-performance}
To understand the impact of the constraints on model performance, we compute the following two metrics for both models on the test set: AUC-ROC and the average precision. The AUC-ROC calculates the balance between the true positive rate and the false positive rate, and the average precision is the trade-off between precision and recall.

\subsubsection{Impact on Model Behavior}\label{sec:methods-impact-behavior}
To explore the impact on the model behavior, we analyze the difference between the two models' behavior using various measures of distance between them. Note, when referring to the impact on model behavior in this paper, we mean the impact on the \textit{observed} difference of the behavior, which can be different to the \textit{true} difference, as the distance measures computed are a sample and can therefore also be influenced by factors such as variance. 

Defining such a measure of distance is relatively intuitive for simple models such as linear regressions. For example, for any two linear regression models, Model A and Model B, represented by their vector of coefficients $\boldsymbol{\beta}^A$, $\boldsymbol{\beta}^B\in \mathbb{R}^d$, one measure of distance could be the L2-norm of the two coefficient vectors, $\lVert\boldsymbol{\beta}^A-\boldsymbol{\beta}^B\rVert_2$, with the models displaying a greater difference in model behavior, the larger that quantity is.

However, for XGBoost models, this task becomes more difficult as there are no natural coefficients to extract from the models. Therefore, we analyze the predictions of the two ML models at the patient-level and evaluate their difference by three measures of distance:
\begin{enumerate}
    \item \textit{Prediction Distance:} this measure is the mean absolute difference of the predicted probabilities of survival for patients in the test set. That is, for any two models, Model A and Model B, suppose $\mathbf{p}^A$ and $\mathbf{p}^B$ represent the predicted probabilities for the two models for all patients in the test set respectively. Let $q$ represent the size of the test set. We then compute the prediction distance, $d_{\text{pred}}$, as follows:
    $$d_{\text{pred}} = \frac{1}{q}\sum_{i=1}^q |\mathbf{p}_i^A-\mathbf{p}_i^B|$$
    
    \item \textit{Ranking Distance:} this measure is the rate of disagreement between the models on the ranking of patients in the test set. To compute this, we select every pair of patients in the test set that has different true outcomes and identify whether the two models rank these two patients in the same order using their predicted probabilities of survival. We then compute the fraction of all pairs where the models disagree on their ranking and name this the ranking distance, $d_{\text{rank}}$. This ranking distance has similarities to a (dis)concordance index.
    
    \item \textit{SHAP Distance:} this measure is the mean L1-norm distance of the SHapley Additive exPlanations (SHAP) values \cite{lundberg} for patients in the test set. That is, for any two models, Model A and Model B, suppose $\mathbf{S}^A, \mathbf{S}^B$, are matrices of size $\mathbb{R}^{q \times (m+1)}$, where $q$ is the size of the test set and $m$ is the number of features, that contain the SHAP values for each model and patient in the test set. The additional column is to account for the SHAP baseline value for each model, which we also include in the L1-norm distance. We then compute the SHAP distance, $d_{\text{SHAP}}$, as follows:
    $$d_{\text{SHAP}} = \frac{1}{q}\sum_{i=1}^q \Big(\sum_{j=1}^{m+1} |\mathbf{S}_{i, j}^A - \mathbf{S}_{i, j}^B|\Big)$$
    
    SHAP can be used to visualize the behavior of ML models, by assigning an additive value to each feature that illustrates feature importance, reflecting both the magnitude and direction of a feature's impact on the individual patient's predicted outcome.
\end{enumerate}

\subsection{Towards a Clinical Feedback-Driven Alignment Approach}\label{sec:methods-impact-clinical}
To inform whether a key requirement for a feedback-driven alignment approach can be met in non-generative AI models (i.e., the requirement that clinicians can determine which model is better aligned with clinical experiential learning when only assessing patient-level predictions), we build on our work from the constraint-guided alignment approach. Specifically, we analyze whether and when clinicians can detect differences between the constrained and unconstrained model in a domain where model performance is similar but observed model behavior is different.

\begin{figure}[h]
\centering
\includegraphics[width=0.95\linewidth]{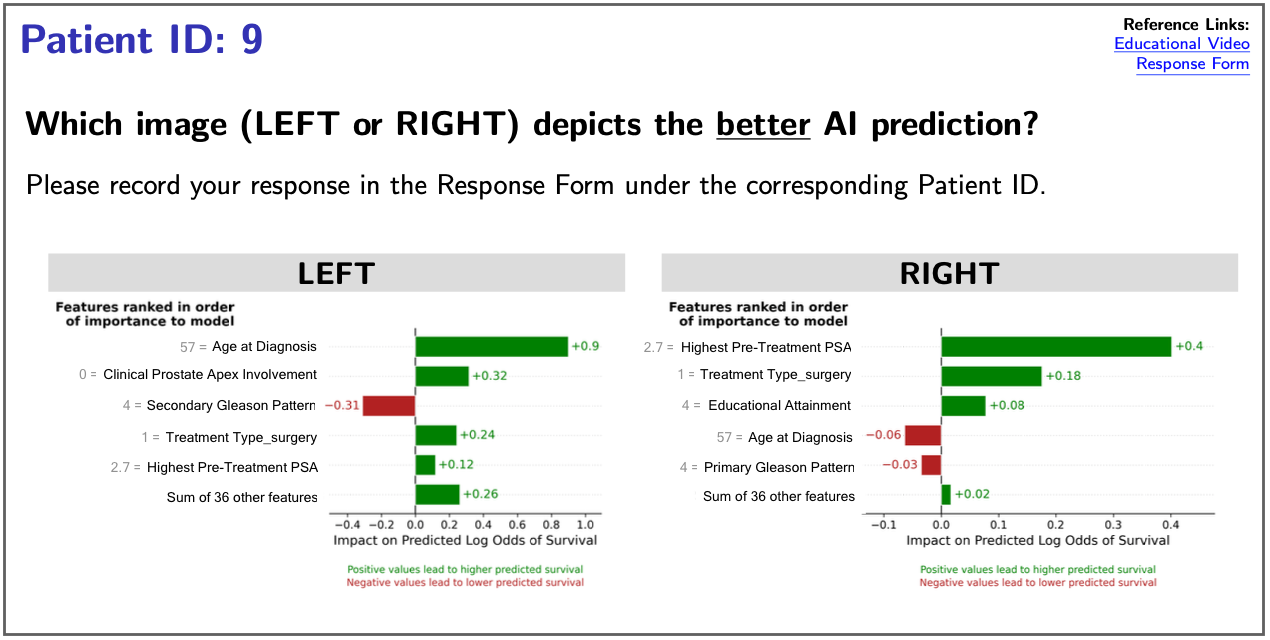}
\caption{Example of a slide provided to a clinician.}
\label{fig:task_slide}
\end{figure}

To research this, we run a small experiment with clinicians. At a high level, the experiment asks clinicians to rank the constrained and unconstrained models for 216 patients in total (36 patients per clinician), using only select information on how the models make their prediction at a patient-level. That is, for the same patient, clinicians receive SHAP bar plots from the two models and are asked to select which one they believe depicts the better AI prediction. Figure \ref{fig:task_slide} shows an illustrative example of a slide that a clinician receives (for patient privacy reasons, feature values have been changed in all examples shown in this paper and therefore do not represent a real patient). The plots provide information on the significance of the features, showing the top five most important features to each model and their values, as well as the magnitude and direction of the impact on the outcome of survival.

Overall, six clinicians participated in the experiment. We were limited to this number of clinicians given general time constraints of clinicians. When asked to participate in our experiment, clinicians are told that our study explores how to align AI prediction models with clinical experiential learning but are provided with no information on the mechanics of how the models shown to them differ. 

To complete the experiment, clinicians follow these three steps sequentially (Appendix \ref{sec:appendix-experiment} provides further details for each step): 
\begin{enumerate}
    \item \textit{Educational Video:} Clinicians first watch a 10-minute video that describes the task to them in more depth. The video includes information on how to complete the task and on how to interpret and compare the images shown to them for the models.
    \item \textit{Study Alignment Form:} After clinicians watch the video, they complete a study alignment form to ensure that they understand the task correctly. The form consists of both true/false questions and multiple-choice questions that test a clinician's comprehension of the data used and SHAP bar plots. In case a clinician gets a question incorrect, we provide them with a detailed explanation of their incorrect answer and ask them to review this before continuing to the task.
    \item \textit{Task:} And finally, clinicians complete the task, each ranking the models for 36 patients (Figure \ref{fig:task_slide} shows an example of the task for one patient), recording both their response and confidence level for each of their selections on a scale from 1-5, with a 1 signifying low confidence and 5 signifying high confidence that their chosen AI prediction is superior to the other.
\end{enumerate}

To select the models and patients we use for the experiment, we first run the constrained and unconstrained model 150 times, varying the seed to generate the train set, but keeping the train set size constant. We then randomly select nine pairs of models (each pair consisting of the constrained and unconstrained model trained for that train set) that have a SHAP distance in the top 25th quantile of SHAP distances across all runs. Then, for each pair, we calculate the L1-norm of the SHAP values for patients in the test set, and select 24 patients such that half are randomly selected and the other half are randomly selected from the top 25th quantile. This provides 216 patients sampled (24 patients from each of the nine model pairs). Lastly, we randomly assign 36 patients to each clinician, giving an equal number from each of the nine model pairs across clinicians, randomizing the order shown on the slides (i.e., whether the constrained model's SHAP bar plot shows up on the `LEFT' or `RIGHT' of the slides).

To understand whether and when clinicians detect differences between the constrained and unconstrained model using our methodology of SHAP plots to elicit preferences, we analyze the relationship between the fraction of times that clinicians chose the constrained versus the unconstrained model and the SHAP distance for the patients assessed through a logistic regression. The regression considers whether a clinician chose the constrained model or not as the dependent variable, and the SHAP distance for the patient assessed as the independent variable of interest. We control for the train set seed (i.e., which of the pairs of models the predictions for the specific patient were generated with), encoded as a categorical feature, as the model-specific fixed effects.

\section{Case Study: Results}
\subsection{Illustration of Inconsistencies and Impact of Constraint-Guided Approach on Univariate Partial Dependence Plots}\label{sec:results-univariate}
\begin{figure}[H]
\centering
\includegraphics[width=\linewidth]{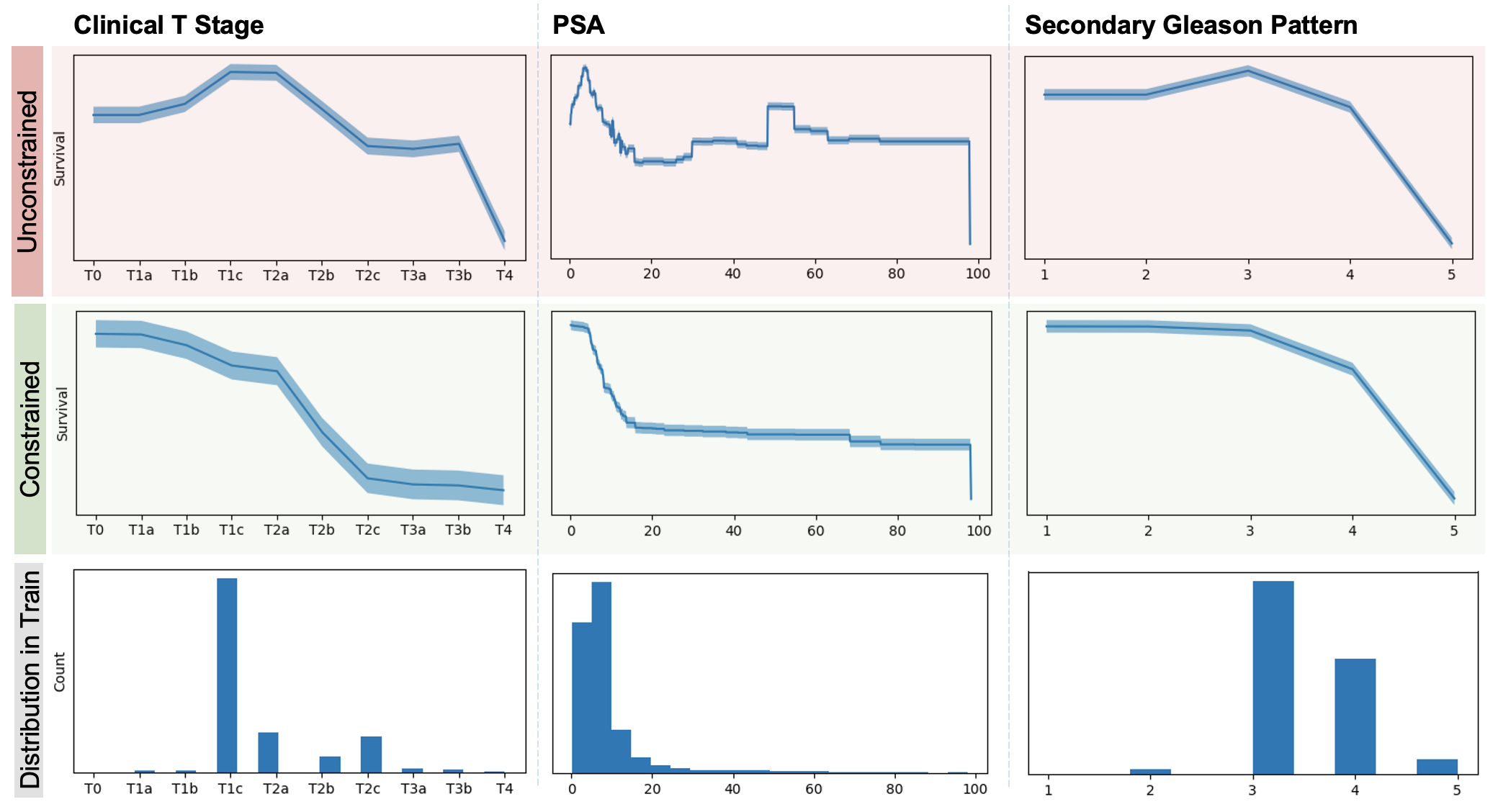}
\caption{Example of univariate partial dependence plots for three key clinical features for a single model run.}
\label{fig:univariate_plot}
\end{figure}

Illustrating an example of the inconsistencies, Figure \ref{fig:univariate_plot} shows the univariate partial dependence plots for an unconstrained model (AUC-ROC: 0.77) trained using the entire 80\% train data ($n=68,345$) for three key clinical features: the stage of the original tumor (clinical T stage), prostate-specific antigen level (PSA), and secondary Gleason pattern which describes the differentiation of cancer cells. According to clinical experiential learning, each of these features should exhibit a monotonically decreasing relationship with the outcome of survival, shown on the y-axis of these graphs, but the model instead predicts a non-monotonic, at times sporadic, behavior, despite the large train set size.

To understand the impact of the constraint-guided alignment, the second row of Figure \ref{fig:univariate_plot} shows the univariate partial dependence plots for the constrained model (AUC-ROC: 0.77) for these three features. Comparing the behaviors between the constrained and unconstrained row of Figure \ref{fig:univariate_plot}, the two models' behaviors are substantially different: the constrained model smooths out the non-monotonic behavior that the unconstrained model exhibits, and is now aligned with the opinions of the clinicians, thereby working as expected. Assessing the behavior between the univariate partial dependence plots and the distribution of the feature values shown in the third row of Figure \ref{fig:univariate_plot}, the inconsistencies occur both in regions of the feature value where the train data is sparse and dense.

\subsection{Impact of Constraint-Guided Approach across Degrees of Underspecification}\label{sec:results-curves}

\subsubsection{Comparison of Model Performance}\label{sec:results-performance}
\begin{figure}[H]
 \centering
 \begin{subfigure}[b]{0.49\textwidth} 
    \centering
   \includegraphics[width=\linewidth]{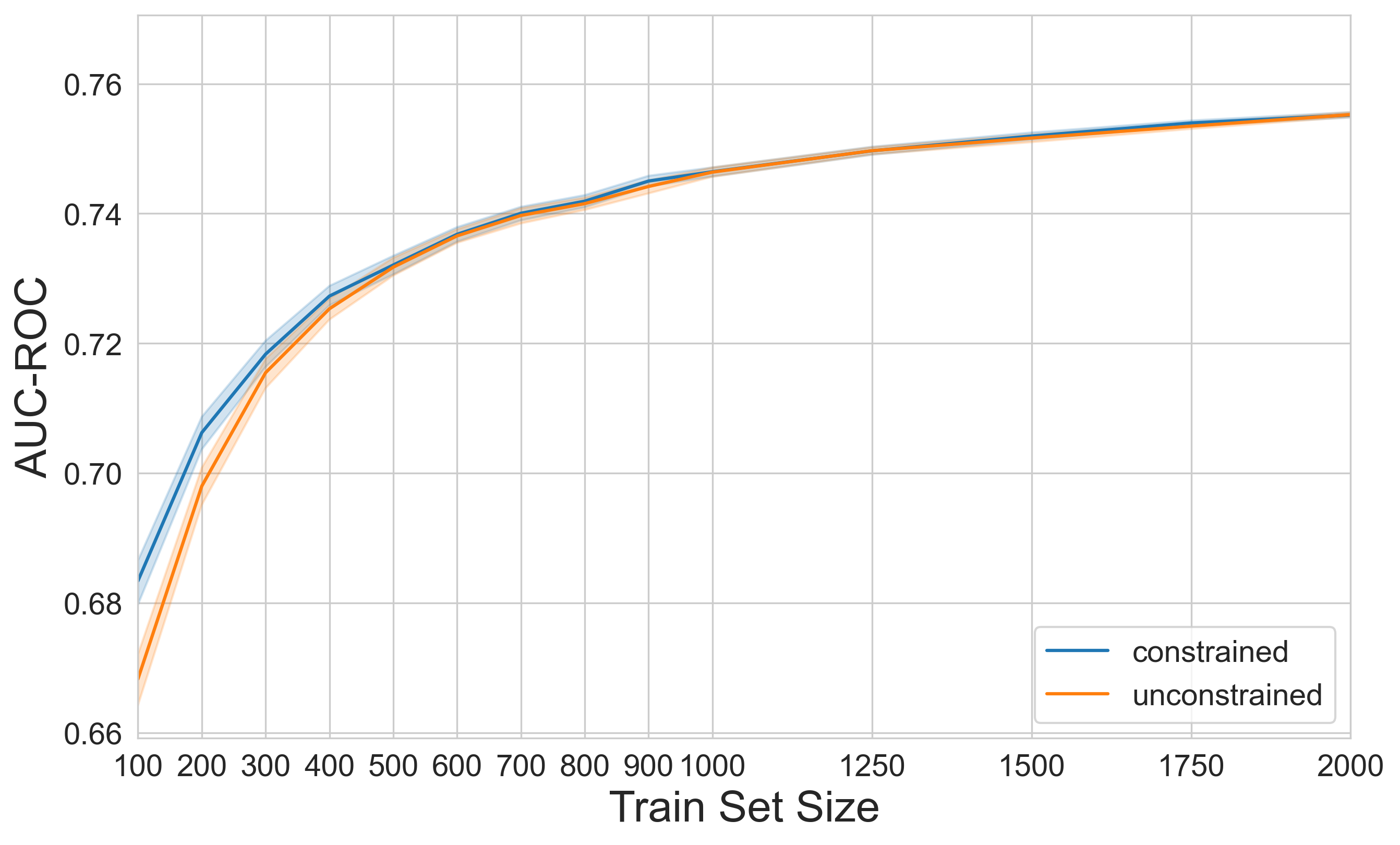}
   \caption{AUC-ROC}
   \label{subfig-1:auc_zoom}
 \end{subfigure}%
 \hfill
 \begin{subfigure}[b]{0.49\textwidth}
    \centering
   \includegraphics[width=\linewidth]{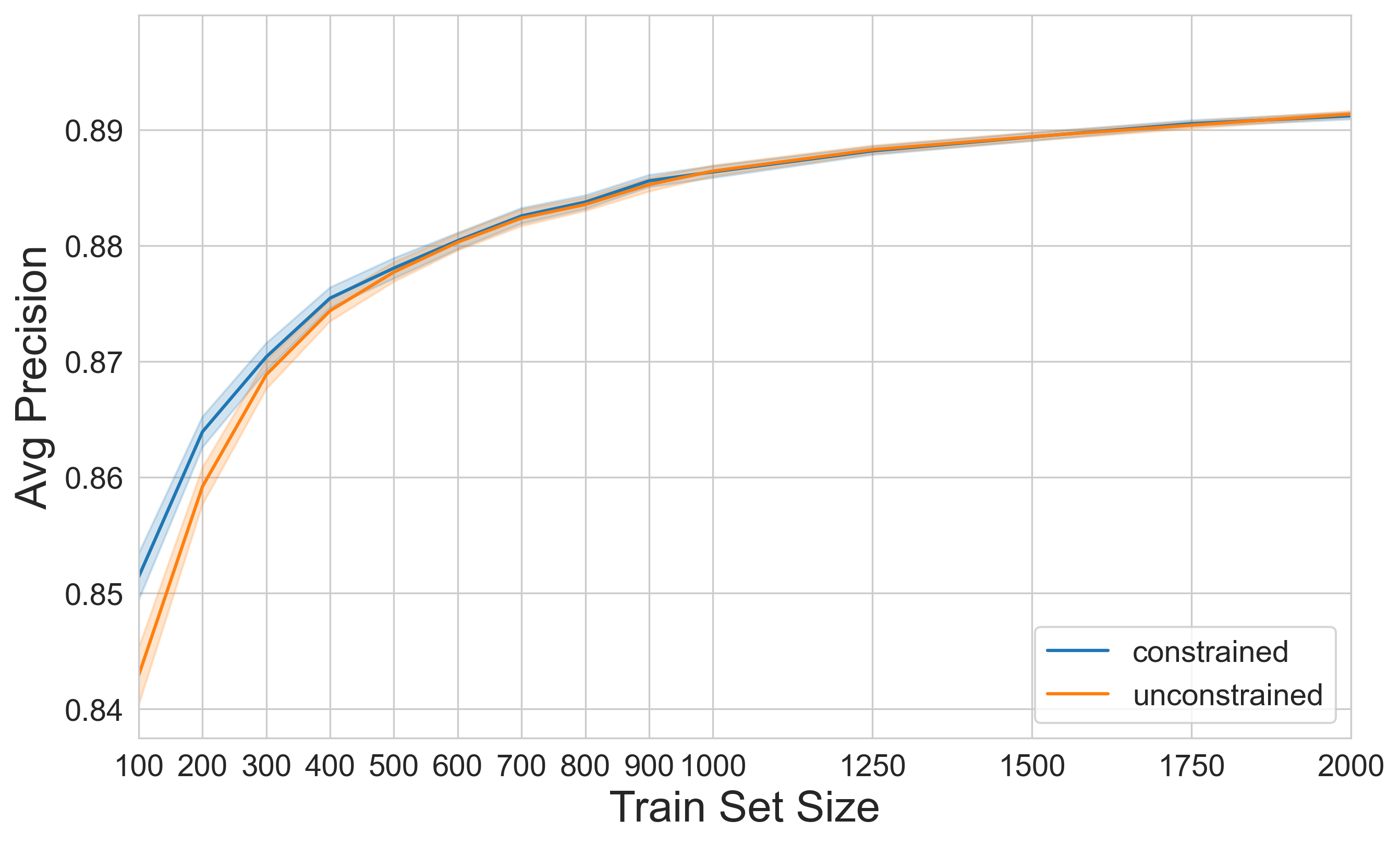}
   \caption{Average Precision}
   \label{subfig-2:prc_zoom}
 \end{subfigure}
 \caption{Learning curves for the constrained and unconstrained model across \textit{train set sizes in the range of [100, 2000]}.}
 \label{fig:performance_zoom}
\end{figure}

\begin{figure}[H]
 \centering
 \begin{subfigure}[b]{0.49\textwidth} 
    \centering
   \includegraphics[width=\linewidth]{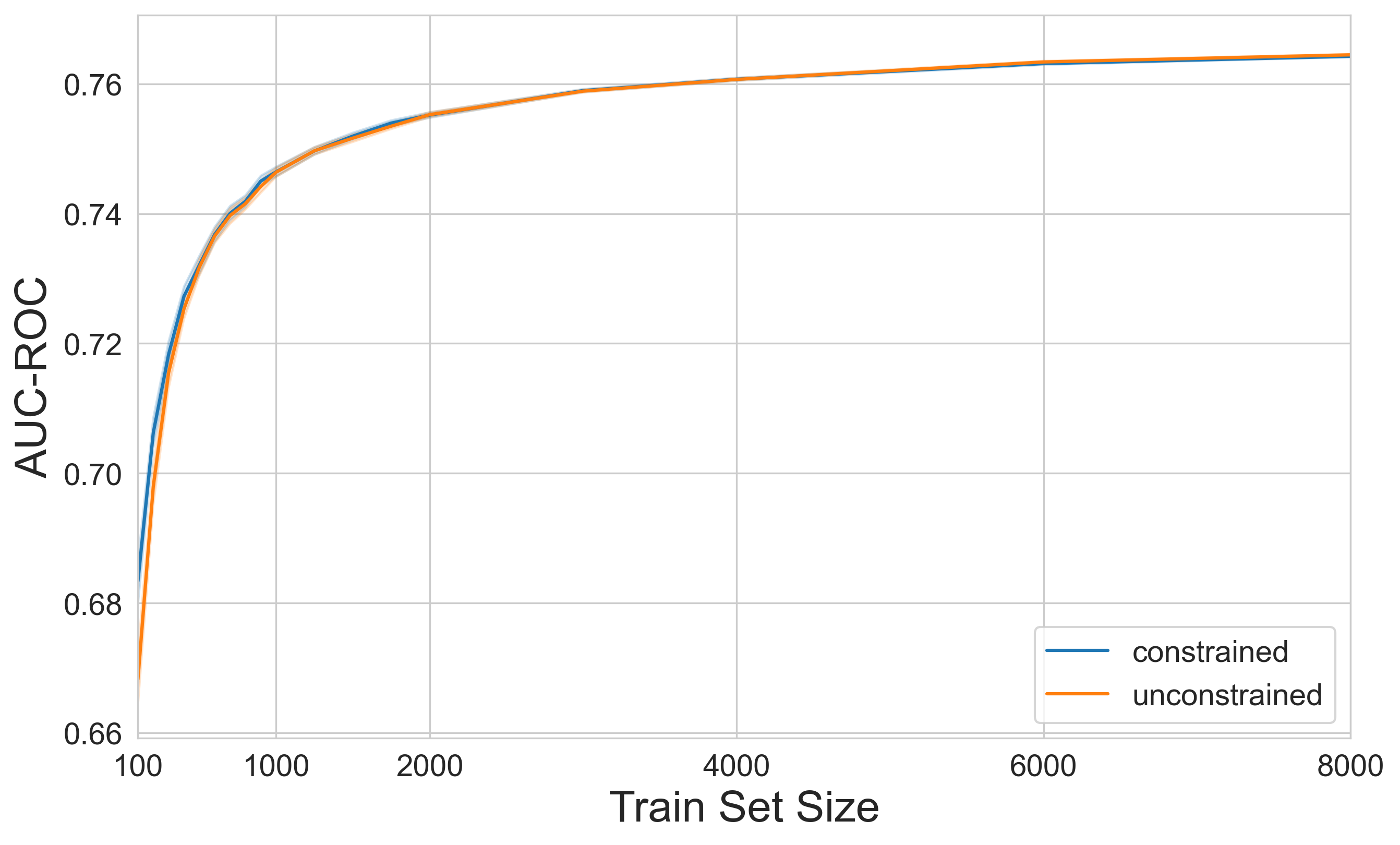}
   \caption{AUC-ROC}
   \label{subfig-1:auc}
 \end{subfigure}%
 \hfill
 \begin{subfigure}[b]{0.49\textwidth}
    \centering
   \includegraphics[width=\linewidth]{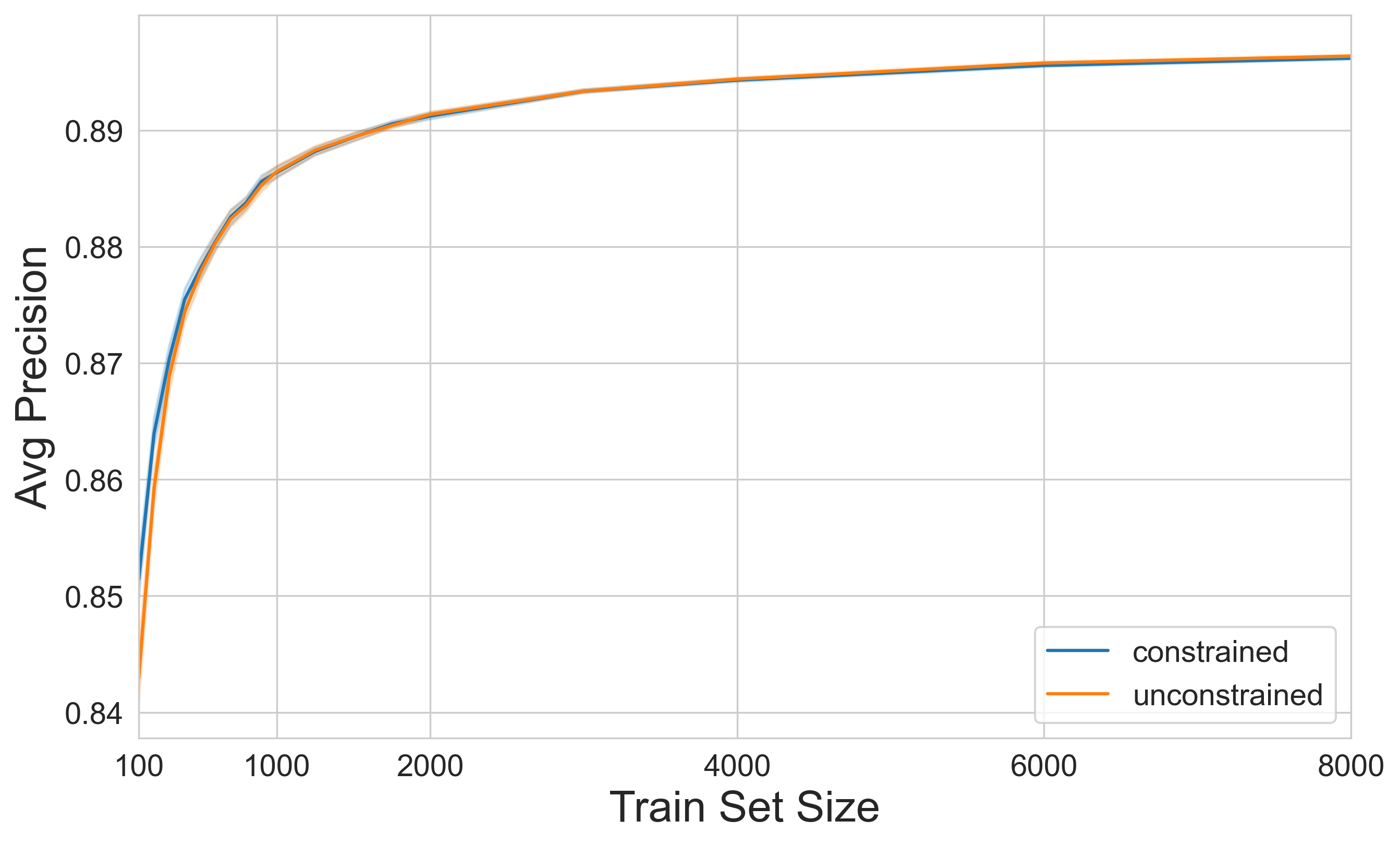}
   \caption{Average Precision}
   \label{subfig-2:prc}
 \end{subfigure}
 \caption{Learning curves for the constrained and unconstrained model across \textit{all train set sizes run}.}
 \label{fig:performance}
\end{figure}

Figure \ref{subfig-1:auc_zoom} and Figure \ref{subfig-2:prc_zoom} plot the mean AUC-ROC and average precision with 95\% confidence intervals across train set sizes, zooming into the range of [100, 2000]. Comparing the curve for the unconstrained model (orange) to the curve of the constrained model (blue), the two models have different performances in very small train set sizes (for train set sizes of $\{100, 200\}$), with the constrained model's curve higher than that of the unconstrained model's for both measures of performance. This is aligned with the idea that prior knowledge in small train set environments can bring benefits. However, as the train set size increases, the curves of the two models overlap, signifying similar levels of performance, and remain overlapping for all train set sizes run (see Figure \ref{fig:performance}).  These results illustrate that, from a performance perspective, the incorporation of the clinically informed constraints does not harm the model performance across all degrees of underspecification tried.



\subsubsection{Comparison of Model Behavior}\label{sec:results-model-behavior}
\begin{figure}[H]
 \centering
 \begin{subfigure}[b]{0.49\textwidth} 
    \centering
   \includegraphics[width=\linewidth]{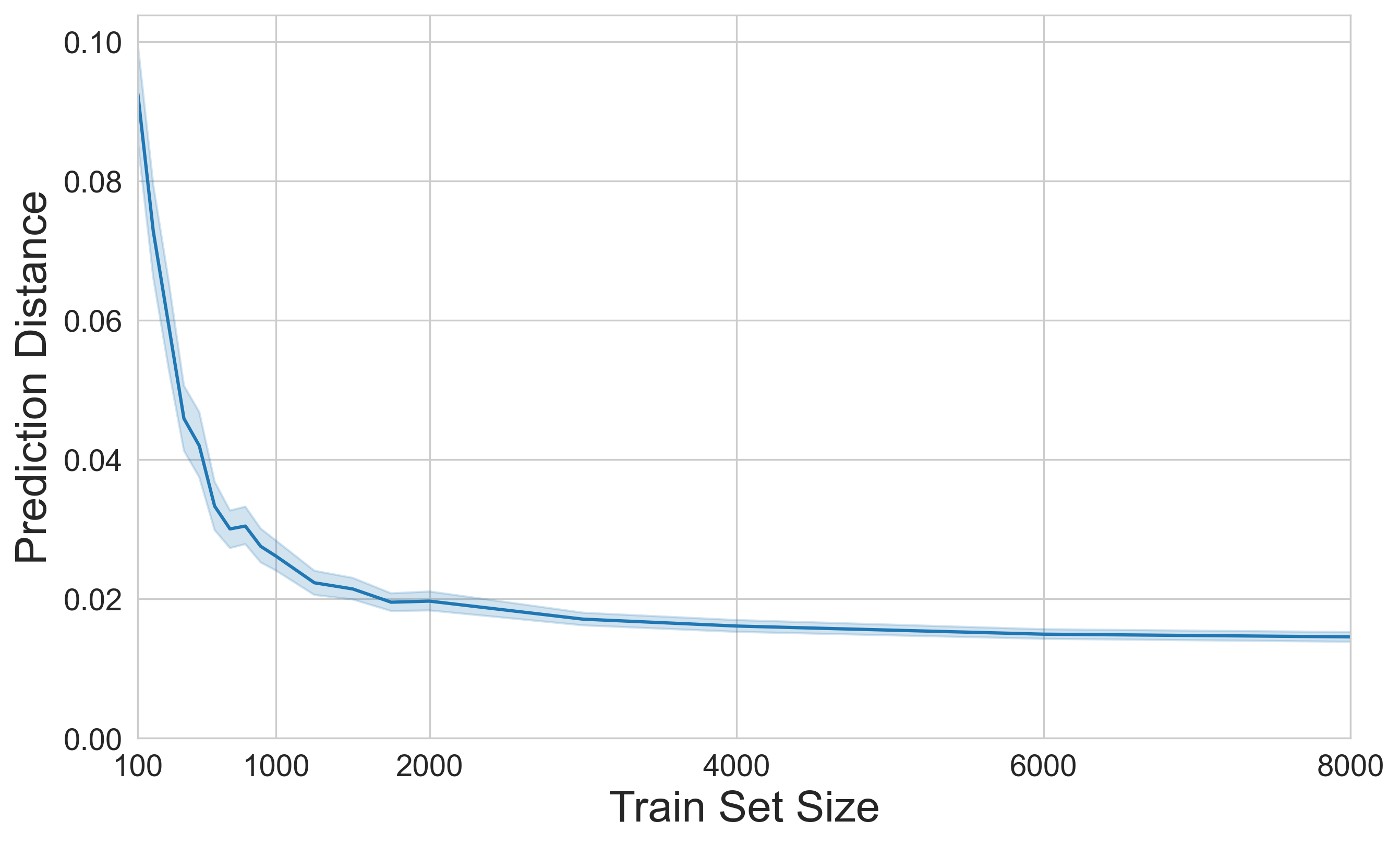}
   \caption{Prediction distance, $d_{\text{pred}}$}
   \label{subfig-1:prediction}
 \end{subfigure}%
 \hfill
 \begin{subfigure}[b]{0.49\textwidth}
    \centering
   \includegraphics[width=\linewidth]{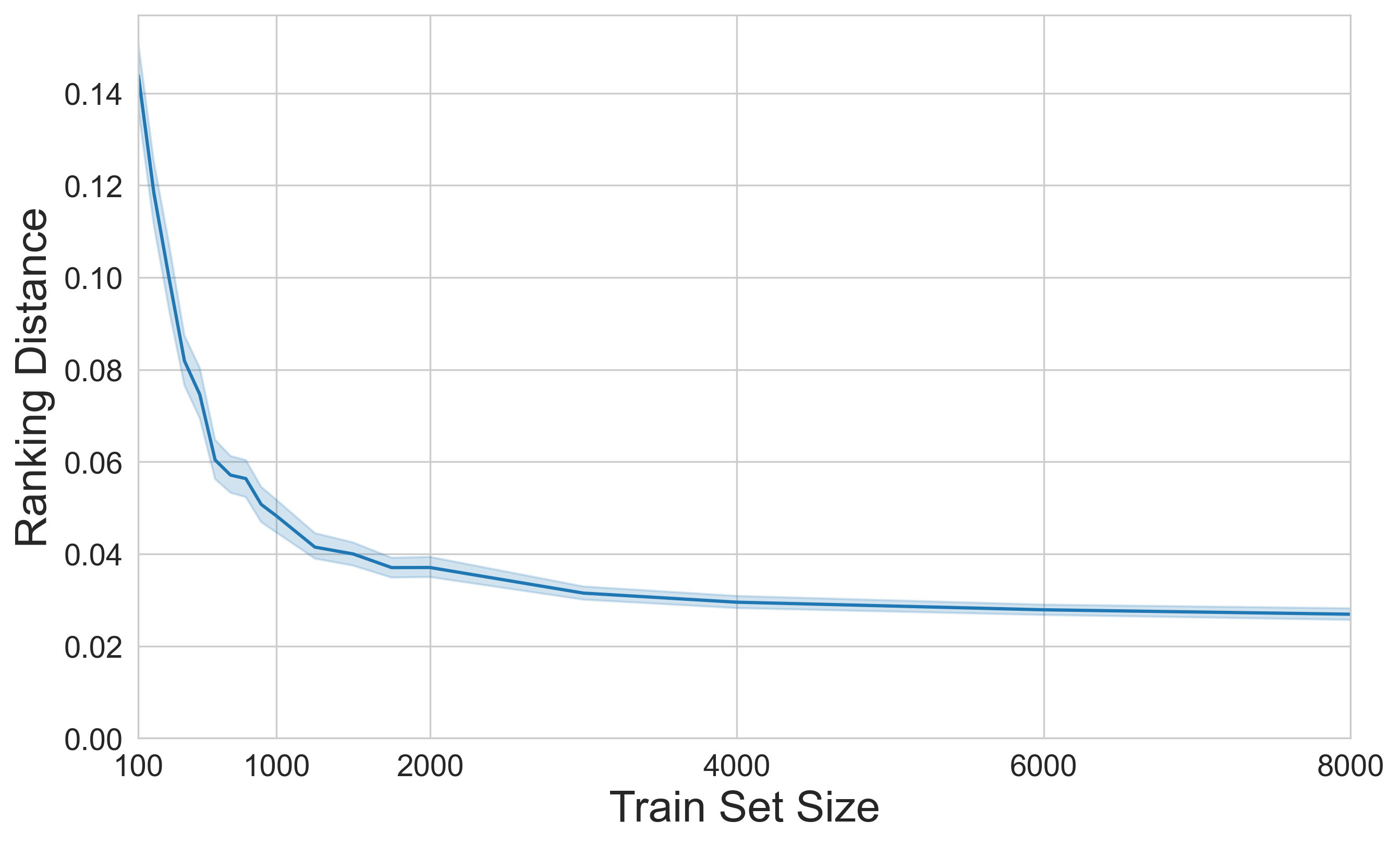}
   \caption{Ranking distance, $d_{\text{rank}}$}
   \label{subfig-2:ranking}
 \end{subfigure}
 \hfill
 \begin{subfigure}[b]{0.49\textwidth}
    \centering
   \includegraphics[width=\linewidth]{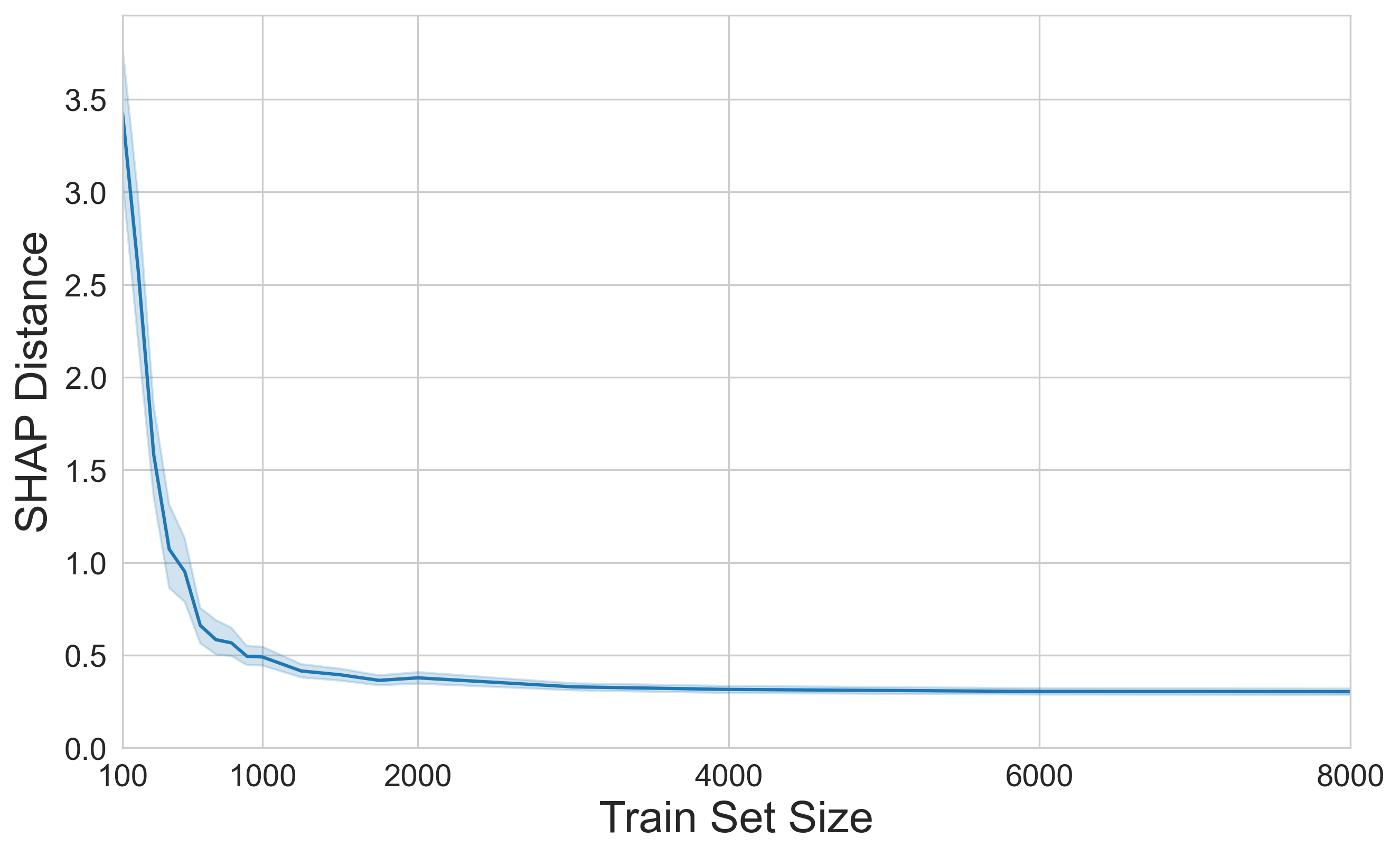}
   \caption{SHAP distance, $d_{\text{SHAP}}$}
   \label{subfig-3:shap_tree}
 \end{subfigure}
 \caption{Distance curves for three measures of distance.}
 \label{fig:distance_curves}
\end{figure}

Figure \ref{fig:distance_curves} plots the three measures of distance (prediction distance, ranking distance, and SHAP distance) defined in Section \ref{sec:methods-impact-behavior}, across different train set sizes. Each curve plots the mean and 95\% confidence interval.

Each of these curves follows a similar behavior: the measure of distance between the models is highest in very small train set sizes (relative to the values in larger train set sizes), then decreases as the train set size increases, and ultimately reaches a plateau in very large train set sizes. While our dataset size limits the maximum train set size that we can use to generate these curves to ensure meaningful interpretations, this non-vanishing fixed distance is also reflected in the univariate partial dependence plots, such as those in Section \ref{sec:results-univariate} produced with a train set size of nearly 70,000. 

While we do not know the exact reason for the slope of the curve, we hypothesize that it is potentially driven by two factors: the true difference of the two models' behaviors and the variance in the difference. In smaller train set sizes, the variance will naturally play a larger role, because the models have greater degrees of freedom and can therefore, by chance, exhibit exceptionally large distances. In larger train set sizes where the distance curves reach a plateau, the variance will likely play a smaller role such that the true distance between the two models is likely displayed. 

Nonetheless, we observe that underspecified models can exhibit different behaviors while maintaining similar performance. One potential explanation for this from the literature might be that underspecified models have inductive biases, which might not impact performance, but can lead to behavior that is misaligned with application-specific requirements (in our case, clinical experiential learning) \cite{d'amour}. These results also suggest that there may be a degree of freedom to better align the model through reliable domain knowledge, without compromising performance. The key to this takeaway is the need for \textit{reliable} domain knowledge. For example, we find that if the constraints imposed are against clinical experiential learning, then the performance is harmed (details in Appendix \ref{sec:appendix-opposite}).


\subsection{Towards a Clinical Feedback-Driven Alignment Approach: Evaluating Clinical Interpretation of Model Behaviors}\label{sec:results-experiment}
For the experiment, we choose a train set size of $n=400$, which corresponds to a domain where on average, the constrained and unconstrained models have similar performance, but different observed behavior. Appendix \ref{sec:appendix-experiment-results} lists summary statistics of the experiment results aggregated across clinicians.

We analyze the logistic regression that considers whether a clinician chose the constrained model as the dependent variable and the SHAP distance as the independent variable, controlling for model-specific fixed effects. Table \ref{tab:experiment_logistic} shows the main result. We find a significant positive coefficient, in log odds, of 0.2947 ($p= 0.031$) for SHAP distance. We find this result to be robust to the regression used, with tests performed using a linear regression and other link functions (e.g., probit, log-log, complementary log-log).

\begin{table}[ht]
\centering
\renewcommand{\arraystretch}{1.4}
\begin{tabular}{ccc}
\hline
\textit{Variable} & \textit{Estimate} & \textit{SE} \\
\hline
SHAP distance & \textbf{0.2947}* & 0.137 \\
{[}Model fixed effects{]} & & \\
\hline
& & \small{* $p<0.05, n = 216$}
\end{tabular}
\caption{Logistic regression coefficients for choosing the constrained model.}
\label{tab:experiment_logistic}
\end{table}

This illustrates that the greater the distance between the two models for a patient, the more accessible it seems for clinicians to select the constrained model, using our methodology of SHAP plots for eliciting preferences between models. These results thereby provide first evidence that one of the key requirements for a feedback-driven alignment approach can be met in non-generative AI models, supporting a potential future opportunity to use user feedback to align such models that suffer from underspecification.

\section{Discussion}
This work showcases examples of model behaviors of modern ML pipelines that are inconsistent with clinical experiential learning in a prostate cancer case study. We show how one can integrate reliable clinical knowledge through clinically informed monotonic constraints into an underspecified ML pipeline to better align it with such expertise. Overall, we find that, across degrees of underspecification controlled by train set size, this modification leads to a non-vanishing difference in the observed model behavior compared to that of an unconstrained model, while maintaining the model performance. Through our randomized experiment with clinicians, we find that the greater the distance between the constrained and unconstrained models for a patient, the more accessible it is for clinicians to select the constrained model, confirming the feasibility of a key requirement for a potential feedback-driven alignment approach. 

Our analyses add nuance to the question of why these inconsistencies may occur in modern ML pipelines. Overman et al. \cite{overman2024aligning} show that as the train set size increases, more complex models can describe advanced non-linear relationships while maintaining high generalizability, making them susceptible to inconsistent behavior in the presence of small noise bias in the data. This perspective combined with our analyses underscores the need for high-quality data and the proactive incorporation of practitioner knowledge to align models with clinical experiential learning, because despite use of the latest ML models and access to large amounts of train data, the models may exhibit inconsistent behavior that is not necessarily reflected by a reduction in model performance. Beyond noise bias, our results in the case study (models with same performance, but different behavior) are also consistent with the hypothesis that an underspecified pipeline may lead to a model that has optimal test set performance, but is misaligned with the end user.

Our results in the case study suggest that there is a degree of freedom to modify an underspecified ML pipeline to correct for misbehavior that goes against domain knowledge to ensure a more trustworthy and, thus, successful deployment of a model in clinical practice. This is in agreement with the claim by D'Amour et al. \cite{d'amour} that in an underspecified domain, the introduction of application-specific constraints should not lead to a compromise between greater reliability of the model in practice and its test performance. An area of future work would be to better understand this degree of freedom by studying whether all alignments are equal, and subsequently their impact on model outcomes and consequences for healthcare stakeholders (e.g., the impact across subgroups of the population).

Inspired by alignment approaches for generative AI models, our randomized experiment illustrates that with our particular methodology for eliciting preferences between models, the larger the degree of disagreement between the unconstrained and constrained models for a patient, the better clinicians can differentiate between them. These results illustrate that one of the key requirements for a feedback-driven alignment approach can be met, thereby supporting a future opportunity to use user feedback to align non-generative AI models that suffer from underspecification. Additional research efforts should focus on how to implement a complete clinical feedback-driven alignment approach into non-generative AI models. Results of the post-experiment survey illustrate that clinicians are very keen on wanting to understand the intricacies of modern ML models, and new tools, such as the SHAP bar plots used in our experiment, can easily be communicated to clinicians.

One limitation of this work is its application of these analyses to a single case study and the potential impact of the data selection bias on our takeaways. The completion of our analyses on only one case study is in part due to the time-intensive process required to collect reliable domain knowledge via strong collaborations between engineering and clinical teams. Creating a library of generally accepted clinical intuition (e.g., which features have a monotonic relationship) that can be readily used as knowledge integrated into ML models is an interesting area of future work. The framework of our survey can provide a first step in achieving this goal.

A further limitation is with respect to the experiment. The experiment considers the responses from only six clinicians, because we require participants with prominent expertise on prostate cancer and who have sufficient time to diligently complete all steps of the experiment. However, due to this sample size, each patient selected is only shown to one clinician. This limits our understanding on the variability of the responses for a specific patient. Furthermore, all models used are trained using a train set size of 400. It would be interesting to investigate the impact on our results as underspecification levels vary. To maximize the power of the study, we were also limited to using a single explainability method to illustrate the difference in the models' feature attributions to clinicians. We chose SHAP bar plots to represent how models make predictions given their recent popularity, interpretability, robustness, and the seamless integration of the SHAP package with the Python Scikit-learn pipeline for ML models. Nonetheless, SHAP bar plots might not fully capture the differences between the constrained and the unconstrained model in a clinician-interpretable and clinically-translatable fashion. Despite the small sample size though, we believe a randomized experiment like ours where domain experts directly evaluate the internal behavior of ML models through an ML explainability interface is rare and offers interesting directions for future research opportunities.

In summary, with the rapidly expanding use of complex ML models in healthcare contexts, it is becoming imperative to develop methods that allow us to gain a deeper understanding of the practical implications of underspecification in modern ML pipelines and the impact of strategies used to mitigate them. Our work introduces a reproducible framework to study these aspects, illustrating a constraint-guided alignment approach and first evidence for meeting a key requirement for a clinical feedback-driven alignment approach. Our analyses highlight the need for both high-quality data and the proactive incorporation of practitioner knowledge to help combat inconsistent behavior with clinical experiential learning.

\bibliographystyle{vancouver}
\bibliography{references}

\appendix
\setcounter{table}{0} 
\renewcommand{\thetable}{\thesection.\arabic{table}}  
\renewcommand\thefigure{\thesection.\arabic{figure}}

\section*{Appendix}

\section{Data Cleaning}

\subsection{Patient Cohort Filters}\label{sec:appendix-patient-cohort}  
To identify the patient cohort, we first apply the following high-level filters: 
\begin{enumerate}[itemsep=-3pt]
    \item The tumor histology is adenocarcinoma
    \item A patient's age is at least 40 years
    \item All patients have only a single malignant primary tumor
    \item Part/all of treatment was performed at the reporting facility
    \item Whether a patient received a prostatectomy is known, as well as the time of the prostatectomy if treated via prostatectomy
    \item Whether a patient received radiation therapy is known, as well as the time of radiation therapy, type of radiation therapy, and location of radiation therapy if treated via radiation therapy
    \item No palliative care was given
    \item All diagnosis dates are on or after a facility's reference date
    \item All cancer treatment was able to be coded in the pre-specified treatment fields established by NCDB
    \item An individual was staged and the TNM-staging edition number is known and categorized as a valid number
    \item The time between the last contact or death and date of diagnosis is known
    \item The vital status (alive or dead) is known of the patient at the time of last contact
\end{enumerate}

Table \ref{tab:all_data_cohort} replicates some of the summary statistics listed in Table \ref{tab:patient_cohort} for our patient cohort, when using the original dataset without any filters imposed. 

\begin{table}[H]
\caption{Summary statistics of original dataset.}
\centerline{\begin{tabular}{  l l l } 
    \cline{2-3} & \\[-1.5ex]
     & \multicolumn{2}{l}{\textbf{Metric for Numerical Features}} \vspace{3pt} \\
    \cline{2-3} & \\[-1.5ex]
     & \textbf{Mean (SD)} & \textbf{Median} \vspace{3pt}\\ 
    \hline\hline & \\[-1.5ex]
    Age at Diagnosis & 65.3 (9.1) & 65.0 \vspace{3pt}\\ 
    Clinical Stage Group & 2.0 (0.7) & 2.0 \vspace{3pt}\\
    Comorbidity Score & 0.22 (0.5) & 0.0 \vspace{3pt}\\
    Highest Pre-Treatment PSA & 13.1 (20.0) &  6.3 \vspace{3pt}\\
    Primary Gleason Pattern & 3.3 (0.5) & 3.0 \vspace{3pt}\\
    Secondary Gleason Pattern & 3.5 (0.6) & 3.0 \vspace{3pt}\\
    \hline & \\[-1.5ex]
    Race = White & 81.3\% & \vspace{3pt}\\
    Insurance Status & & \vspace{3pt}\\
    \hspace{0.5cm} Medicaid & 2.6\% & \\
    \hspace{0.5cm} Medicare & 45.6\% & \\
    \hspace{0.5cm} Other Gov & 1.8\% & \\
    \hspace{0.5cm} Private & 46.1\% & \\
    \hspace{0.5cm} Uninsured & 1.7\% & \vspace{3pt}\\
    \hline\hline
\end{tabular}}
\label{tab:all_data_cohort}
\end{table}

\subsection{Features with Missing Values}\label{sec:appendix-features}  
In the case study, we primarily use ML models that can handle missing values, such that we do not impute missing values before training the models. Table \ref{tab:missing_values} outlines the features with missing values and the fraction of time they are missing for each. 

\begin{table}[H]
\centering
\caption{List of features with missing values and the fraction of time they are missing.}
\label{tab:missing_values}
\begin{tabular}{|l|r|}
\hline
\textbf{Feature} & \textbf{Fraction Missing} \\
\hline
Clinical Prostate Apex Involvement & 0.32 \\
Clinical T Stage & 0.30 \\
Clinical Stage Group & 0.25 \\
Spanish or Hispanic Origin & 0.10 \\
Highest Pre-Treatment PSA & 0.09 \\
Secondary Gleason Pattern & 0.08 \\
Primary Gleason Pattern & 0.08 \\
Insurance Status & 0.03 \\
Rurality and Urban Influence & 0.03 \\
Educational Attainment & 0.03 \\
Median Household Income & 0.03 \\
Race & 0.02  \\
\hline
\end{tabular}
\label{tab:missing-fraction}
\end{table}

\section{Details for Experiment with Clinicians}\label{sec:appendix-experiment}
As discussed in Section \ref{sec:methods-impact-clinical}, in order for clinicians to complete the experiment, they had to follow three steps sequentially: 1) watch an educational video; 2) complete a study alignment form; and 3) complete the task. The sections below provide example snapshots of each of these for a better understanding.

\subsection{Educational Video}
\begin{figure}[H]
\centering
\begin{subfigure}[b]{0.7\textwidth} 
   \centering
  \includegraphics[width=\linewidth]{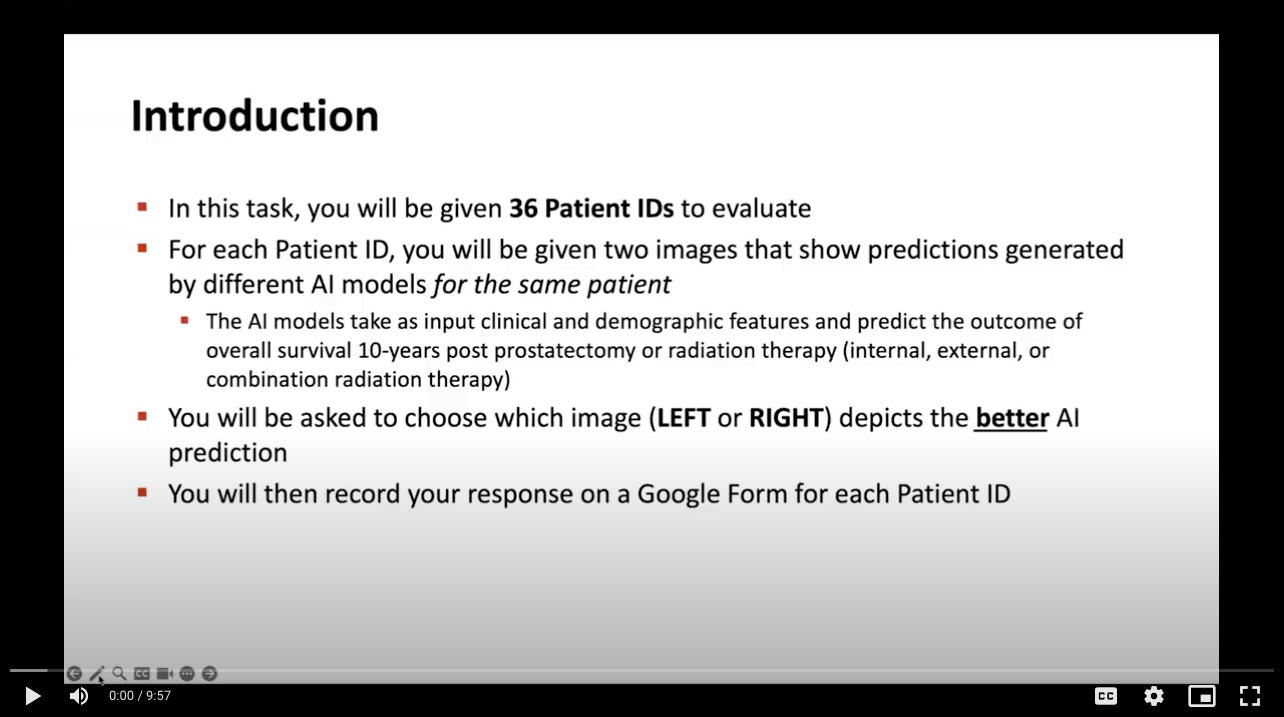}
  \label{subfig-1:video1}
\end{subfigure}
\hfill
\begin{subfigure}[b]{0.7\textwidth}
\centering
\includegraphics[width=\linewidth]{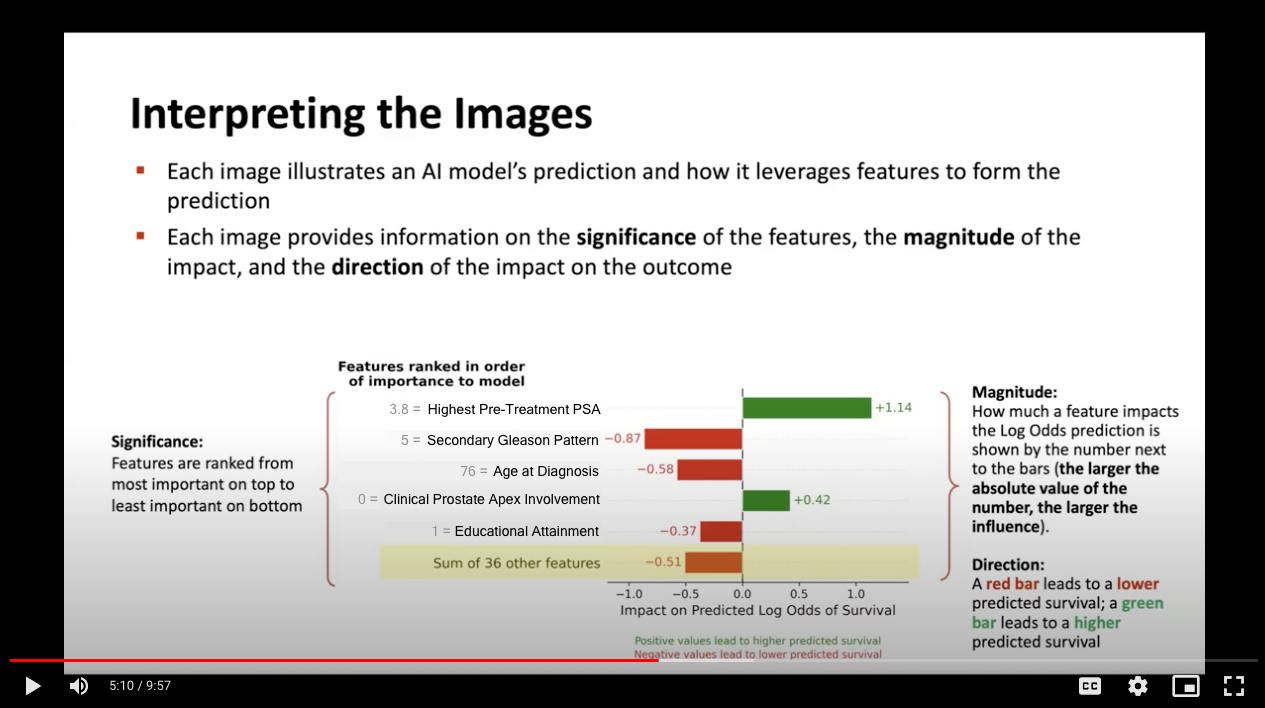}
\label{subfig-2:video2}
\end{subfigure}
\caption{Snapshots from the educational video.}
\label{fig:educational_video}
\end{figure}

Figure \ref{fig:educational_video} provides snapshots of the educational video, showing the introduction slide and an example of how the SHAP bar plots are explained to clinicians. 

\subsection{Study Alignment Form}
\begin{figure}[H]
\centering
\begin{subfigure}[b]{0.49\textwidth} 
   \centering
  \includegraphics[width=\linewidth]{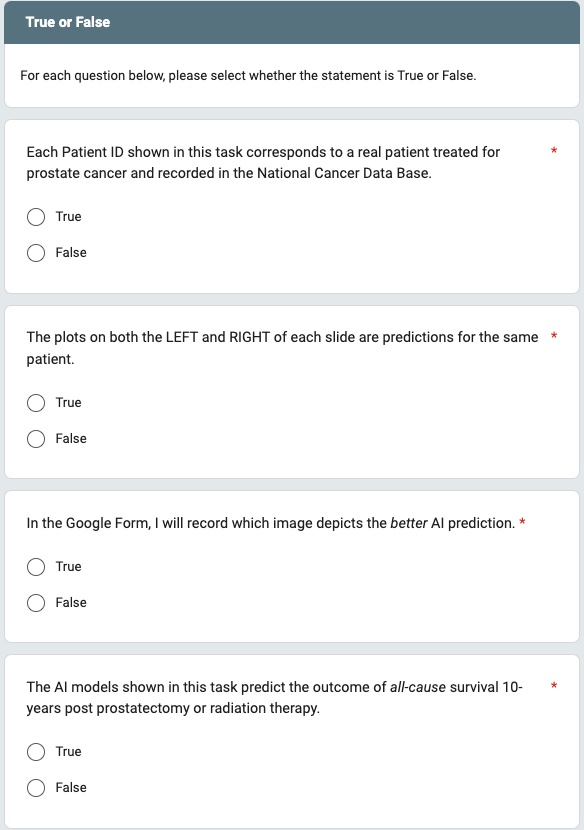}
  \caption{True/false questions}
  \label{subfig-1:survey1}
\end{subfigure}
\hfill
\begin{subfigure}[b]{0.49\textwidth}
\centering
\includegraphics[width=\linewidth]{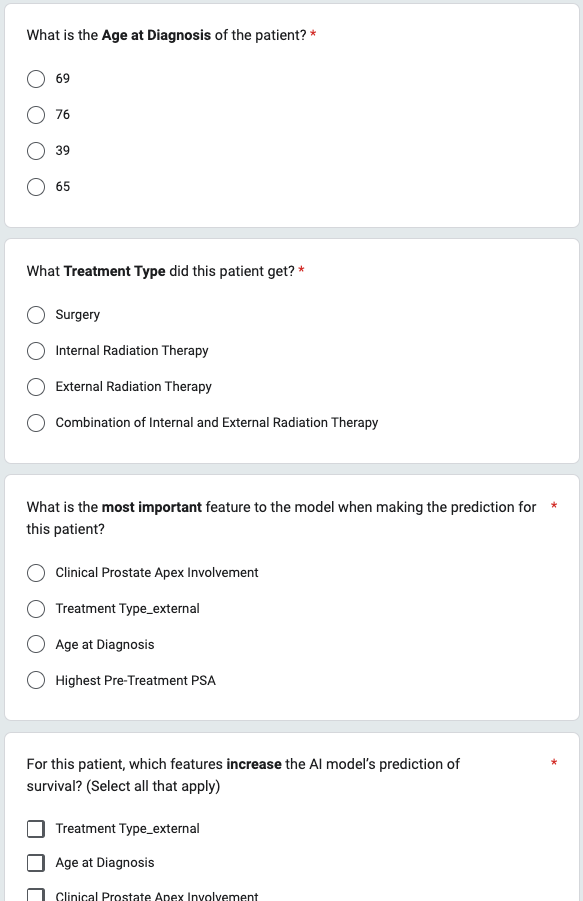}
\caption{Multiple choice questions}
\label{subfig-2:survey2}
\end{subfigure}
\caption{Snapshot of the study alignment form.}
\label{fig:survey}
\end{figure}

The study alignment form ensures that we can test whether a clinician understands the task before they begin. The form consists of both true/false questions and multiple choice questions. Figure \ref{subfig-1:survey1} shows the true/false questions asked in the form. For the multiple choice questions, we provide clinicians with an example SHAP bar plot, and ask them questions such as those in Figure \ref{subfig-2:survey2}, testing their ability to interpret SHAP bar plots. Overall, the form has 10 questions (4 true/false, 6 multiple choice).

\subsection{Task}
\begin{figure}[H]
\centering
\includegraphics[width=0.6\linewidth]{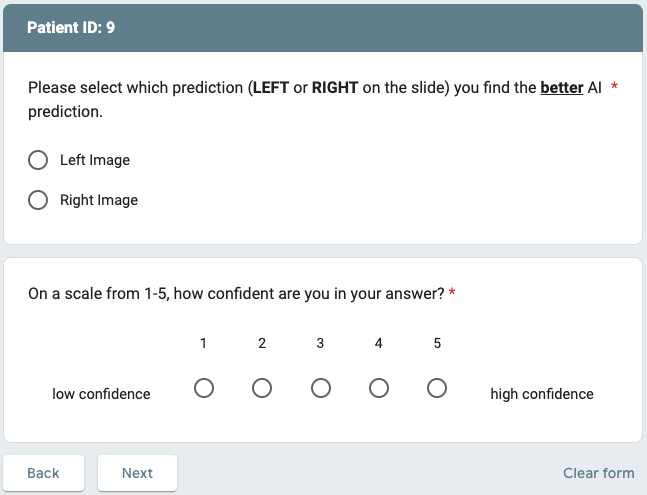}
\caption{Snapshot of the response form used by clinicians to record their answers.}
\label{fig:response_form}
\end{figure}

For the task, we provide clinicians with a file of the 36 patients for them to evaluate and a link to a response form to record their response. Figure \ref{fig:response_form} shows an example of a snapshot of the response form. We also provide clinicians with a features explanations page that lists all features that the models have access to and a brief explanation of each feature (i.e., a general description and the values the features can take on).

\section{Degree of Freedom or Reliable Domain Knowledge?}\label{sec:appendix-opposite}
Our main findings show that the constrained and unconstrained models have similar levels of performance for all train set sizes (besides very small train set sizes). This raises the question: is there just a large degree freedom in constraining underspecified ML pipelines such that one could add arbitrary, or even unintuitive, constraints without hurting the performance, or are we restricted to constraints elicited from reliable domain knowledge? 

\begin{figure}[H]
 \centering
 \begin{subfigure}[b]{0.49\textwidth} 
    \centering
   \includegraphics[width=\linewidth]{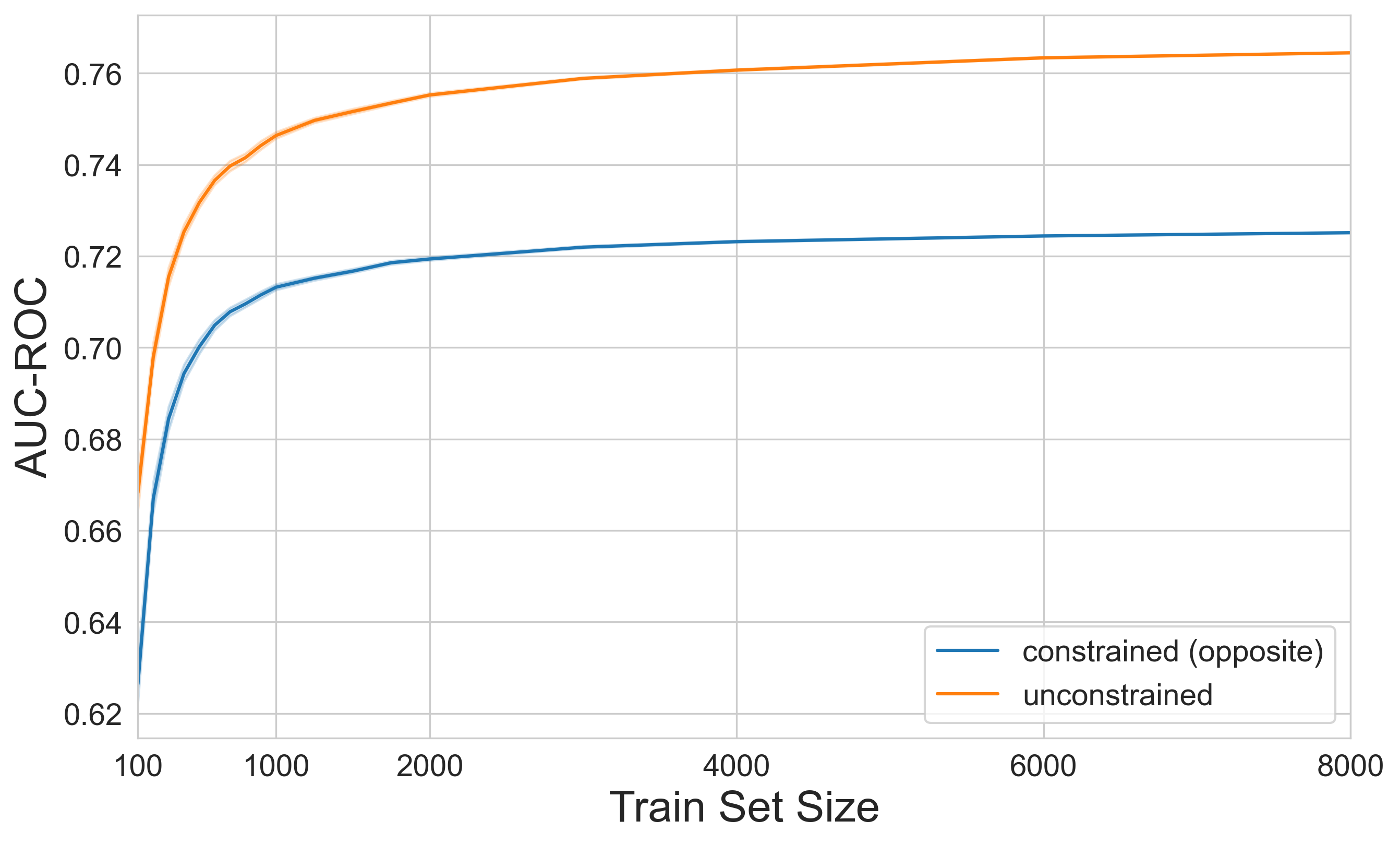}
   \caption{AUC-ROC}
   \label{subfig-1:auc_opposite}
 \end{subfigure}%
 \hfill
 \begin{subfigure}[b]{0.49\textwidth}
    \centering
   \includegraphics[width=\linewidth]{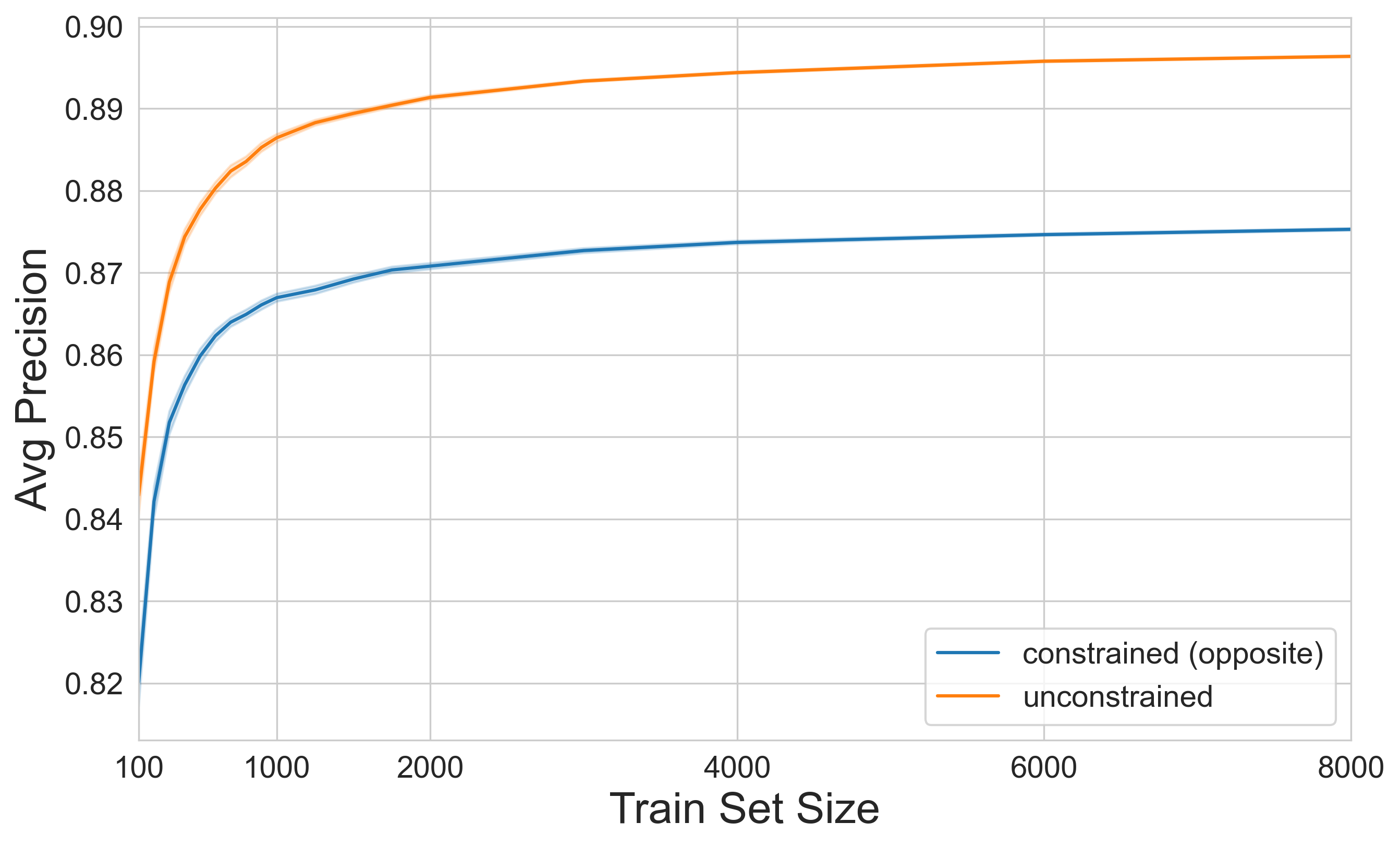}
   \caption{Average Precision}
   \label{subfig-2:prc_opposite}
 \end{subfigure}
 \caption{Measures of performance for the constrained model with constraints in opposite direction and unconstrained model across train set sizes.}
 \label{fig:performance_opposite}
\end{figure}

To test this, we constrain the constrained features in the constrained model in the opposite direction, such that the features are forced to have a monotonically increasing relationship with the outcome of survival. These constraints therefore go against clinical experiential learning. The unconstrained model is not impacted. Similar to our other analyses, we then compare the performance of the two models across train set sizes, for 300 runs of the XGBoost models for each train set size. Figure \ref{fig:performance_opposite} shows the AUC-ROC and the average precision for the changed constrained model in blue for different train set sizes. In this scenario, the changed constrained model performs significantly worse compared to the unconstrained model across all train set sizes and thereby levels of underspecification, highlighting that we do not have absolute freedom in constraining a model. 

\section{Summary Statistics of the Experiment}\label{sec:appendix-experiment-results}
Table \ref{tab:experiment_results} lists summary statistics of the experiment results aggregated across clinicians.

\begin{table}[H]
\caption{Summary statistics for experiment results, with 95\% CIs.}
\centerline{\begin{tabular}{  l l }  
    \hline\hline & \\[-1.5ex]
    \% of time the constrained model was chosen & 47.7\% (41.2-54.2)\vspace{3pt}\\ 
    SHAP Distance (Mean) & \vspace{3pt}\\
    \hspace{0.5cm} Unconstrained model chosen & 4.88 (4.25-5.57)\\
    \hspace{0.5cm} Constrained model chosen & 5.42 (4.64-6.26)\vspace{3pt}\\
    Confidence Level (Mean) & \vspace{3pt}\\
    \hspace{0.5cm} Unconstrained model chosen & 2.65 (2.45-2.89)\\
    \hspace{0.5cm} Constrained model chosen & 2.62 (2.38-2.86)\vspace{3pt}\\
    \hline\hline
\end{tabular}}
\label{tab:experiment_results}
\end{table}


\end{document}